\pdfoutput=1

\documentclass[11pt]{article}

\usepackage[preprint]{acl}

\usepackage{times}
\usepackage{latexsym}

\usepackage[T1]{fontenc}

\usepackage[utf8]{inputenc}

\usepackage{microtype}

\usepackage{inconsolata}

\usepackage{graphicx}

\usepackage{url}
\usepackage{latexsym}
\usepackage{tikz}
\usepackage[utf8]{inputenc}
\usepackage{xcolor}
\usepackage{tcolorbox}
\usepackage{enumitem} 
\usepackage{amsmath}
\usepackage{multirow}
\usepackage{adjustbox}
\usepackage{graphicx}
\usepackage{subcaption}
\usepackage{hyperref}
\usepackage{array}

\title{FaMTEB: Massive Text Embedding Benchmark in Persian Language}

\author{%
  Erfan Zinvandi$^{\ddagger\mathsection}$, Morteza Alikhani$^{\ddagger\mathsection}$, Mehran Sarmadi$^{\ddagger\mathsection}$,  Zahra Pourbahman$^\mathsection$ \\
   \textbf{Sepehr Arvin}$^\ddagger$\textbf{,} \textbf{Reza Kazemi}$^{\ddagger\mathsection}$\textbf{,} \textbf{Arash Amini}$^{\ddagger\mathsection}$
   \\
   $^\ddagger$MCINext, $^\mathsection$Sharif University of Technology \\
  \texttt{\{e.zeynvandi1376, morteza.alikhani95, mehran.sarmadi99, zahra.pourbahman95, aamini, }\\ \texttt{reza.kazemi\}@sharif.edu, sepehr.arvin@outlook.com}\\}

\begin{document}
\maketitle
\begin{abstract}
In this paper, we introduce a comprehensive benchmark for Persian (Farsi) text embeddings, built upon the Massive Text Embedding Benchmark(MTEB). Our benchmark includes 63 datasets spanning seven different tasks: classification, clustering, pair classification, reranking, retrieval, summary retrieval, and semantic textual similarity. The datasets are formed as a combination of existing, translated, and newly generated data, offering a diverse evaluation framework for Persian language models. Given the increasing use of text embedding models in chatbots, evaluation datasets are becoming inseparable ingredients in chatbot challenges and Retrieval-Augmented Generation systems. As a contribution, we include chatbot evaluation datasets in the MTEB benchmark for the first time. 
In addition, in this paper, we introduce the new task of summary retrieval which is not part of the tasks included in standard MTEB.
Another contribution of this paper is the introduction of a substantial number of new Persian-language NLP datasets suitable for training and evaluation, some of which have no previous counterparts in Persian.
 We evaluate the performance of several Persian and multilingual embedding models in a range of tasks. This work introduces an open-source benchmark with datasets, code and a public leaderboard~\footnote{Persian space on \url{https://huggingface.co/spaces/mteb/leaderboard}}.
\end{abstract}

\section{Introduction}

Text-embedding models aim to generate a semantic vector representation of text, which is helpful in addressing various natural language processing (NLP) tasks such as clustering, classification, semantic textual similarity (STS), information retrieval (IR), and more~\citep{lewis2020retrieval}. 

To evaluate the performance of models on these tasks, most existing benchmarks are task-specific and fail to assess model capabilities across multiple tasks. For instance, an information retrieval model like Dense Passage Retrieval (DPR)~\citep{karpukhin2020dense} may perform well on retrieval tasks but fails to achieve satisfactory results in find semantic textual similarity (STS). To address this limitation, the massive text embedding benchmark (MTEB)~\citep{muennighoff-etal-2023-mteb} introduced a comprehensive evaluation suite across eight diverse NLP tasks, effectively meeting the evaluation needs of text-embedding models for the English language. However, since the primary focus of this benchmark is on English, it does not adequately assess the performance of models in low-resource languages like Persian.

In this work, we introduce \textbf{FaMTEB}, a large-scale Persian benchmark for evaluating Persian text-embedding models, enabling users to select the most suitable text-embedding model for their specific tasks. This benchmark comprises of $63$ datasets across $7$ tasks, and we compare the performance of $15$ existing Persian or multilingual language models on it. Of these, $24$ datasets were pre-existing in Persian, while we contribute $39$ new datasets. The newly introduced datasets were developed using three distinct approaches: web-based collection ($4$ dataset), translation of existing English datasets ($16$ datasets), and generation through large language models (LLMs) as synthetic datasets ($19$ datasets). The quality of all datasets has been independently assessed to ensure reliability.

To assess the generalizability of text-embedding models across various NLP tasks, it is essential to conduct a comprehensive evaluation on diverse problems. 
Since one of the primary and recently highly popular applications of text-embedding models is in retrieval-augmented generation (RAG) systems and chatbots, 
a portion of the newly curated datasets is dedicated to evaluating these systems. This aspect is explored for the first time within 
this benchmark. 
Needless to say that the Persian language is among the low-resource languages in the web and gathering relevant and high quality data is not readily available. 

Below we highlight the main contributions of this work:
\begin{itemize}
\item Introduction of FaMTEB, a large-scale Persian benchmark for evaluating Persian
text-embedding models,
\item introduction of a significant number of new Persian datasets in the field of NLP 
suitable for training and evaluation, some of which have no prior equivalents, 
\item introduction of the new task of \emph{summary retrieval}, which was not among the $8$ tasks included in MTEB,
\item introduction of several datasets related to chatbot challenges and RAG systems, which have been included in the MTEB benchmark for the first time.
\end{itemize}

\section{Related work}
\subsection{Benchmarks}
Before the advent of the MTEB\citep{muennighoff-etal-2023-mteb}, the evaluation of text embeddings was fragmented across various task-specific and domain-specific benchmarks. Early efforts like the semantic textual similarity benchmarks, including STS-Benchmark~\citep{cer-etal-2017-semeval} and SICK~\citep{marelli-etal-2014-sick} datasets, primarily assessed embeddings for their ability to capture semantic relationships between text pairs. While effective in measuring specific aspects, 
these benchmarks were narrow in scope, focusing on small datasets and failing to represent real-world diversity.

The GLUE~\citep{wang-etal-2018-glue} Benchmark broadened the evaluation landscape by incorporating tasks such as natural language inference, sentiment analysis, and sentence similarity. However, the benchmark is dedicated exclusively to the English language, limiting its applicability to multilingual NLU research and offering limited insight into the generalization capabilities of raw embeddings. Similarly, information retrieval benchmarks like MS MARCO~\citep{DBLP:journals/corr/NguyenRSGTMD16} evaluate embeddings for domain-specific applications, such as search engine optimization. While these benchmarks are invaluable for retrieval tasks, they do not generalize 
effectively across a wide range of embedding evaluation use-cases.

The fragmented nature of these benchmarks underscore several limitations. First, task diversity is insufficient; most benchmarks aim at specific applications such as similarity, classification, or retrieval without covering clustering or zero-shot classification. 
Second, inconsistencies in evaluation protocols and metrics make it challenging to compare results across models. Finally, existing benchmarks are not designed for scalability or extensibility; this complicates incorporating 
new datasets or tasks as embedding techniques evolve.

These limitations led to the creation of MTEB, a unified and comprehensive benchmark that addresses these gaps. MTEB evaluates text embeddings in a wide range of tasks, including semantic similarity, clustering, classification, retrieval, bitext mining, pair classification, and reranking.

Despite the advancements brought by MTEB, its main focus remains on the English language, creating a need for benchmarks tailored to other languages. Although MTEB supports multilingual evaluation, the representation of certain languages and specific linguistic nuances can be limited. To address this shortcoming, new benchmarks inspired by MTEB have been introduced for languages such as Chinese~\citep{10.1145/3626772.3657878}, Polish~\citep{poswiata2024pl}, and French~\citep{ciancone2024mteb}. These language-specific benchmarks aim to evaluate the quality of text embeddings in tasks and datasets that reflect the unique linguistic and cultural characteristics of these languages, ensuring broader applicability of text embedding models across the global linguistic landscape.

For the Persian language, besides a limited number of evaluation datasets for isolated tasks, no comprehensive evaluation dataset for text embedding models has been introduced yet. 
In the STS task, the Farsick~\citep{9721521} dataset is available, which is a machine-translated version of the SICK dataset. 
For the IR task, the multilingual MIRACL~\citep{zhang-etal-2023-miracl} dataset  supports the Persian language. 

\subsection{Embedding models}
The evolution of text embedding models could be seen as a significant shift from traditional methods like GloVe~\citep{pennington-etal-2014-glove} to more advanced context-sensitive models. GloVe and Word2Vec~\citep{10.5555/2999792.2999959} set the foundation for dense word embeddings by representing words as fixed vectors based on co-occurrence statistics. Although these models worked well to capture word-level semantics, they were limited by their static nature, which did not account for polysemy or contextual meaning.

This limitation was addressed with the introduction of ELMo~\citep{peters-etal-2018-deep} and later BERT~\citep{Devlin2019BERTPO}, which offered contextual embeddings by considering the surrounding text. BERT, leveraging the transformer architecture, became a breakthrough by generating deep bidirectional embeddings, capturing richer contextual information. However, BERT's focus on token-level embeddings necessitated fine-tuning for specific tasks, which could be computationally expensive.

To solve this, sentence transformers, like SBERT~\citep{Reimers2019SentenceBERTSE} were developed to provide high-quality, task-specific sentence embeddings. These models used the transformer architecture with a Siamese network structure to generate embeddings suitable for tasks such as semantic similarity and information retrieval, making them efficient and versatile for sentence-level understanding.

In recent years, open-source models such as BGE~\citep{10.1145/3626772.3657878}, E5~\citep{Wang2022TextEB}, and GTE~\citep{li2023generaltextembeddingsmultistage} have been introduced for text embedding, demonstrating strong performance in tasks like STS, retrieval, and clustering in English. In the Persian language, various foundational models such as ParsBERT~\citep{Farahani2020ParsBERTTM}, FaBERT~\citep{Masumi2024FaBERTPB}, and TookaBERT~\citep{sadraeijavaheri2024tookabertstepforwardpersian} have been released, which are capable of understanding textual information well; however, they cannot be directly applied to a wide range of tasks. To date, models that perform well in Persian text embedding tasks mainly consist of multilingual networks such as BGE-m3~\citep{chen-etal-2024-m3}, mE5~\citep{wang2024multilinguale5textembeddings}, and OpenAI's text embedding models text-embedding-3 and text-embedding-4.

\section{The Persian MTEB}

\begin{figure*}[t]
\includegraphics[width=\textwidth]{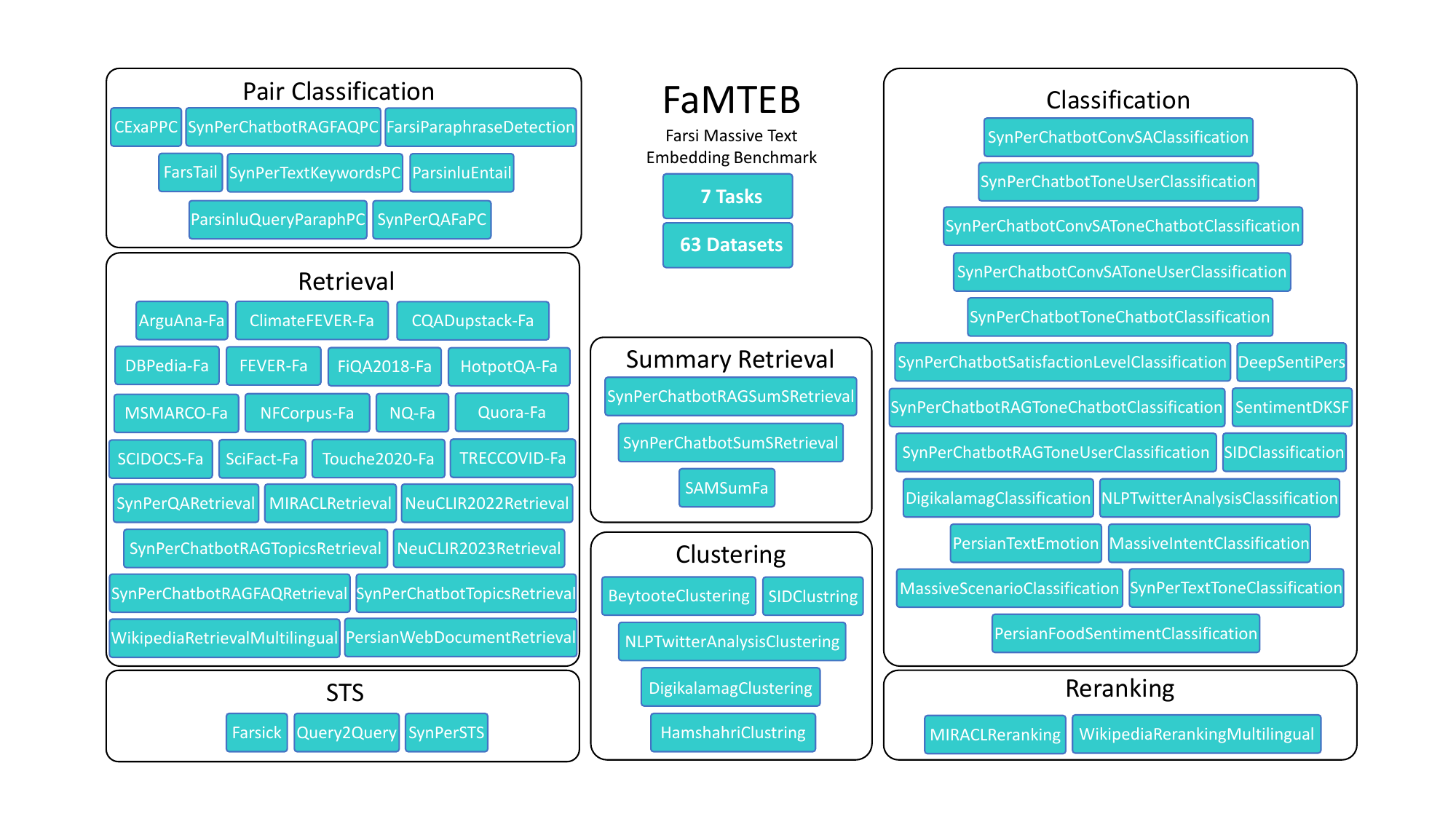}
\centering
\caption{An overview of the FaMTEB evaluation dataset.}
\label{fig:famteb}
\end{figure*}

\subsection{Task definition and evaluation strategy}
This benchmark consists of  $7$ tasks: classification, clustering, semantic textual similarity, retrieval, reranking, pair classification, and summary retrieval. The evaluation methods for all tasks, except summary retrieval, which will be explained later, are detailed in the MTEB~\citep{muennighoff-etal-2023-mteb}.

\textbf{Summary Retrieval:} in this task, we have two input sets. The first set contains a collection of texts, while the second set contains text summaries from the first set. In this evaluation, we search for the text vector from the first set within the second set using the cosine distance. The first retrieved result must be the summary of the corresponding text. F1 score is used as the primary metric for summary retrieval, with accuracy, precision, and recall also being computed to assess the performance.

\subsection{New datasets}
\subsubsection{BEIR-Fa}
\label{BEIR}
A significant portion of our retrieval evaluation data consists of the BEIR benchmark~\citep{Thakur2021BEIRAH} datasets, which have been translated into Persian using the Google Translate service. Similarly, the mMARCO~\citep{DBLP:journals/corr/abs-2108-13897} evaluation data is also derived by translating the MARCO~\citep{DBLP:journals/corr/NguyenRSGTMD16} dataset using this service. Since the maximum input length for Google Translate is 5,000 characters, documents exceeding this length were truncated. We call this newly generated dataset, BEIR-Fa. Further details are provided in  Appendix~\ref{sec:data_evaluation}.

\subsubsection{Synthetic Persian STS}
\label{SynPerSTS}
To the best of our knowledge, no independently constructed dataset exists for the STS task in Persian. Available datasets are typically derived from translations of English datasets or created using unsupervised methods. For instance, the FarSICK dataset is a translation of the SICK dataset. To address this lack of datasets in Persian, we propose constructing a new dataset using a LLM.

The first step involves compiling a collection of sentences with a suitable distribution of diverse topics. To achieve this, we sample data from various websites covering topics such as Wiki, news, articles, informal, and others. For generating each sample, the documents from each site are segmented at the sentence level using the \emph{nltk} library. Then, those sentences are selected that meet the following criteria: they must contain between $35$ to $200$ characters in length, and at least $30$ percent of their characters must be in Persian. These conditions ensure that each sentence is complete, meaningful, and not overly ambiguous.

Subsequently, we employ a prompt-based method, 
 inspired by the rules for determining sentence similarity levels  proposed in  SemEval~\citep{cer-etal-2017-semeval}. However, instead of the six levels of similarity used in SemEval, we define five levels, with the final score being a discrete value ranging from $1$ to $5$ (see Tabel \ref{tab:sts_labels}).
Finally, we provide samples with the given prompt to \textit{GPT-4o-mini}\footnote{https://chat.openai.com/} to generate STS data. For more information, check Appendix~\ref{DatasetsDetails}.

\begin{table}[]
    \centering
    \begin{adjustbox}{width=\columnwidth}
    \begin{tabular}{|c|l|}
        \hline
         1 & The two sentences are completely dissimilar but the writing\\ & structure of the two texts can be similar. \\
         \hline
         2 & The two sentences are not equivalent but share the same
         topic. \\
         \hline
         3 & 
         The two sentences are roughly equivalent, but some important\\ & information differs/missing. \\
         \hline
         4 & The two sentences are mostly equivalent, but some unimportant\\ & details differ \\
         \hline
         5 & The two sentences are completely equivalent, as they mean the\\ & same thing. \\
         \hline
    \end{tabular}
    \end{adjustbox}
    \caption{Similarity score definition based on \protect\citep{cer-etal-2017-semeval}.}
    \label{tab:sts_labels}
\end{table}

\subsubsection{Synthetic Persian chatbot}
Given the increasing prevalence and usage of diverse chatbots, along with the various challenges in chatbot systems that text embeddings and language models can address, the creation of datasets specific to chatbot-related challenges has become highly significant.

The goal of constructing the synthetic Persian chatbot dataset is to generate multiple datasets focused on chatbot-related challenges. The dataset generation process involves several steps. First, $175$ distinct topics are identified as conversation subjects for users interacting with chatbots. These topics are selected to be diverse, covering various chatbot types and scenarios, with the final number of samples for each topic determined accordingly.

Five distinct tones are considered for the conversations including formal, casual, childish, aggressive, and street-style. Additionally, $19$ combinations of user and chatbot tones are designed by pairing these tones. Probabilities are assigned to each combination, ensuring that the total probabilities across the combinations equaled one.

Five levels of user satisfaction are defined: very bad, bad, average, good, and excellent. For each topic, samples are generated to correspond to each satisfaction level, with the number of samples for the \lq\lq excellent\rq\rq~level being twice as much as other levels.

For each sample, a tone combination for the user and chatbot is selected based on the assigned probabilities. Each sample included a topic, a user satisfaction level, a user tone, and a chatbot tone. Using these specifications, a prompt is created and provided to \textit{GPT-4o-mini} to generate a conversation between the user and the chatbot. The generated conversation adheres to the specified topic, user tone, chatbot tone, and user satisfaction level. Additionally, the prompt instructs the model to produce supplementary outputs alongside the conversation, such as a summary of the dialogue and the key topics discussed.

Finally, from this comprehensive dataset, several specialized datasets are derived, each tailored for specific purposes, which are detailed in Appendix~\ref{DatasetsDetails}.

\subsubsection{Synthetic Persian chatbot RAG}
Unlike the Synthetic Persian chatbot dataset, where each sample contained a complete conversation between the user and the chatbot, in the synthetic Persian chatbot RAG dataset, the chat is not necessarily complete. During the conversation between the user and the chatbot, one of the user’s messages is considered as the new user input, and the output should respond to this message. This message may either be a follow-up, i.e., it dependents on previous interactions, or it may not be contextually linked to prior conversations, making the dataset particularly suitable for RAG systems.

Each sample consists of the user's new message along the history of previous user-chatbot conversations. This design aligns with the requirements of RAG systems. The dataset shares components with the synthetic Persian chatbot dataset, such as using the same $175$ topics as conversation subjects and applying the $19$ combinations of user and chatbot tones from the previous dataset. However, this dataset does not include the final user satisfaction level.

To construct the dataset, a number from the set $\{0, 2, 4, \dots, 20\}$ is randomly chosen to determine the number of historical messages in the conversation. Each number in the set has an assigned selection probability. Based on the specified parameters, a suitable prompt is crafted and provided to \textit{GPT-4o-mini} to generate the output. This output includes the desired conversation (conversation history plus the new user message), a summary of the conversation, the main topics of the conversation, and two FAQ-related samples. The positive FAQ sample is a question-answer pair that resolved the user's need expressed in the new message, while the negative FAQ sample is a question-answer pair related to the conversation but does not address the user's need in the new message.

The resulting dataset designed with this structure, is further used to create several specialized datasets, each tailored for different applications. These derived datasets are discussed in Appendix~\ref{DatasetsDetails}.

\subsubsection{Synthetic Persian chatbot conversational sentiment analysis}
\label{SynPerChatbotConvSAClassification}
The primary objective of this dataset is to create a sentiment analysis dataset focusing on user emotions during interactions with chatbots. The dataset incorporates nine distinct emotions: anger, satisfaction, friendship, fear, jealousy, surprise, love, sadness, and happiness. Each emotion is assigned a selection probability, with the total probabilities summing to $1$.

Three tones are considered for both the user and the chatbot: formal, casual, and childish. This results in nine possible tone combinations, each with a specific selection probability. Each sample in the dataset includes a topic chosen from the $175$ predefined topics, one of the nine tone combinations for the user and chatbot, one of the nine predefined emotions, and an intensity level for the selected emotion, categorized into three levels: neutral, moderate, and high.

Using these specifications, an appropriate prompt is crafted and provided to \textit{GPT-4o-mini}. The output consistes of a conversation between the user and the chatbot that adheres to the required topic, tone, emotion, and emotion intensity level.

The generated dataset is further utilized to create additional specialized datasets tailored for specific sentiment analysis applications. These derived datasets are elaborated on in Appendix~\ref{DatasetsDetails}.

\subsubsection{Synthetic Persian keywords}
Similar to the synthetic Persian STS, a collection of texts is selected from various curated Persian websites. In this case, the texts are structured as paragraphs rather than single sentences.

Using a well-designed prompt with \textit{GPT-4o-mini}, the desired outputs are generated for each text. These outputs consist of a list of keywords, which represent the key ideas of the text, and a list of non-keywords, which are words in the text that are not considered keywords.

This dataset serves as a comprehensive resource for tasks related to keyword extraction and text analysis in Persian language processing. Further details are provided in Appendix~\ref{DatasetsDetails}.

\subsubsection{Synthetic Persian tone}

We consider the four tones of formal, casual, childish, and literary, and assign each a specific probability. For this dataset, we utilize the paragraphs from the \textit{synthetic Persian keywords dataset}.

For each paragraph, a tone is randomly selected based on the predefined probabilities. Using \textit{GPT-4o-mini}, the paragraph is rewritten in the chosen tone. This process results in the creation of the \textit{synthetic Persian tone dataset}, which provides a valuable resource for studying tone transformation and stylistic variations in Persian text.

\subsubsection{Synthetic Persian QA}

The objective of this dataset is to create a large-scale QA dataset in Persian. To construct this dataset, we select many web pages from curated Persian websites and extract their main content.

For each page, the extracted text is provided to \textit{GPT-4o-mini} along with an appropriately designed prompt. The model generated multiple question-and-answer pairs based on the content of each page. This process resulted in the creation of the \textit{Synthetic Persian QA Dataset}, a valuable resource for question-answering tasks and Persian natural language understanding.

\subsubsection{Query to query}
\label{Query2Query}
Logically, the queries yielding similar results in a search engine are inherently similar. Thus, the degree of overlap in search results between two queries roughly measures their similarity. The data for this dataset is collected using anonymous user queries and search results from the \emph{Zarrebin} search engine\footnote{https://zarebin.ir/}, a platform designed for the Persian language. The similarity score between query pairs is normalized to account for differences in the number of responses. The similarity score is calculated by dividing the intersection of responses by the harmonic mean of the total number of responses for the two queries:

\begin{equation*}
  {\rm score} = \frac{\rm intersection}{\frac{(2*Q_1*Q_2)}{Q_1+Q_2}},
\end{equation*}
where, \lq\lq intersection\rq\rq~represents the number of common responses, while $Q_1$ and $Q_2$ refer to the total number of responses for the first and second queries, respectively.
 
Queries containing harmful or toxic content are filtered for integrity of the dataset. The resulting scores are categorized into three levels: unrelated, partially related, and fully related. Thresholds of $0.2$ and $0.4$ are determined empirically. The dataset is beneficial as it includes query pairs with spelling errors or incomplete sentences, requiring the model to handle deficiencies. Furthermore, the dataset includes queries requiring meaning comprehension, such as when users search for a song in different ways, expecting similar results. This highlights the model's ability to assess semantic relationships despite phrasing differences.

\subsubsection{SID}
\label{SID}
One approach to dataset creation involves leveraging existing data available on the web. For instance, the MTEB clustering dataset is constructed using category labels from the Arxiv website. Similarly, we adopt this method to develop a clustering dataset for Persian articles. By crawling the \textbf{SID}\footnote{https://sid.ir/} website, which hosts a categorized collection of Persian articles, we curate a dataset comprising of $8$ categories. We utilize the title and abstract of each paper, concatenated with two newline characters as input text. Each input text is assigned a single category as the output.

\subsubsection{BeytooteClustering}
\label{BeytooteClustering}
The BeytooteClustering dataset is derived from the collection and categorization of documents from the Beytoote\footnote{https://www.beytoote.com/} website. Beytoote is a Persian blog that publishes posts on a variety of topics. By crawling the pages of this site, we collect $200,000$ documents and classify them into $22$ categories based on the tags present in the URLs. Since each document can be quite lengthy and cover a wide range of topics, we consider only the \lq\lq summary\rq\rq~tag from the website, which provides a condensed version of the document's content, as the main text. In addition, we remove categories containing fewer than $500$ documents to ensure that each category has a sufficient number of entries. The final dataset consists of $95,851$ documents in $19$ categories.

\subsection{Data Collection}
To construct the classification dataset, we utilize existing data related to this task in Persian, such as DigikalamagClassification~\citep{Farahani2020ParsBERTTM}, NLPTwitterAnalysisClassification~\citep{nlp_twitter_analysis}, SentimentDKSF~\citep{hezar2023}, PersianTextEmotion~\citep{persian_text_emotion}, PersianFoodSentimentClassification~\citep{Farahani2020ParsBERTTM}, DeepSentiPers~\citep{PourmostafaRoshanSharami2020DeepSentiPersND},
MassiveIntentClassification~\citep{fitzgerald-etal-2023-massive},
and MassiveScenarioClassification~\citep{fitzgerald-etal-2023-massive}. In addition, we incorporate newly created datasets  SynPerChatbotConvSAClassification, SynPerChatbotToneUserClassification, SynPerChatbotToneChatbotClassification, SynPerChatbotRAGToneUserClassification, SynPerChatbotRAGToneChatbotClassification, SynPerChatbotConvSAToneUserClassification, SynPerChatbotConvSAToneChatbotClassification, SynPerChatbotSatisfactionLevelClassification, and SynPerTextToneClassification, as described in previous sections.

For the clustering dataset, we also employ existing Persian datasets such as HamshahriClustring~\citep{farsi_news}, DigikalamagClustering, and NLPTwitterAnalysisClustering, along with the newly developed datasets BeytooteClustering and SIDClustring, detailed earlier.

The STS dataset is compiled using the same approach, utilizing the existing Persian dataset Farsick~\citep{9721521} in conjunction with the newly introduced datasets SynPerSTS and Query2Query.

Similarly, the pair classification dataset is constructed using the existing Persian datasets CExaPPC~\citep{9786243}, FarsiParaphraseDetection~\citep{farsi_paraphrase_detection}, ParsinluEntail~\citep{khashabi-etal-2021-parsinlu}, ParsinluQueryParaphPC~\citep{khashabi-etal-2021-parsinlu}, and FarsTail~\citep{DBLP:journals/corr/abs-2009-08820}, complemented by the new datasets SynPerChatbotRAGFAQPC, SynPerTextKeywordsPC, and SynPerQAPC, previously discussed.

For the retrieval dataset, we rely on Persian datasets PersianWebDocumentRetrieval\citep{10553090}, NeuCLIR2023Retrieval~\citep{Lawrie2023OverviewOT}, NeuCLIR2022Retrieval~\citep{lawrie2024overviewtrec2023neuclir}, and MIRACL~\citep{zhang-etal-2023-miracl}, along with the newly created datasets BEIR-Fa, SynPerChatbotRAGFAQRetrieval, SynPerChatbotTopicsRetrieval, SynPerChatbotRAGTopicsRetrieval, and SynPerQARetrieval, as elaborated in prior sections.

The reranking dataset was assembled by combining the existing Persian datasets MIRACLReranking and WikipediaRerankingMultilingual~\citep{wikidump}.

Finally, the summary retrieval dataset is prepared using the newly created datasets SynPerChatbotSumSRetrieval, SynPerChatbotRAGSumSRetrieval, and SAMSumFa, as explained in previous sections.

Additional details about these datasets are provided in the appendices.

\subsection{Models}
\label{models}
The models we evaluate are categorized into three main groups. The first group consists of multilingual text embedding models that support Persian, including multilingual-e5~\citep{wang2024multilinguale5textembeddings}, BGE-m3~\citep{chen-etal-2024-m3}, GTE-multilingual~\citep{zhang-etal-2024-mgte}, paraphrase multilingual, paraphrase-multilingual-MiniLM-L12-v2~\citep{reimers-2019-sentence-bert}, LaBSE~\citep{feng-etal-2022-language}, and Jina~\citep{sturua2024jinaembeddingsv3multilingualembeddingstask}. The second group consists of text embedding models specifically trained for Persian, such as BERT-WLNI, RoBERTa-WLNI, sentence-transformer-parsbert, and Tooka-SBERT. The third group includes base model transformers trained in Persian, namely ParsBERT~\citep{Farahani2020ParsBERTTM}, FaBERT~\citep{Masumi2024FaBERTPB}, and TookaBERT~\citep{sadraeijavaheri2024tookabertstepforwardpersian}.

\subsection{Results}
In this section, we evaluate the models introduced in Section \ref{models} on the FaMTEB benchmark. The results of our experiments are presented in Table \ref{tab:faMTEB_results}. It can be observed that the Jina model achieves the highest accuracy on the FaMTEB benchmark in terms of average accuracy. This model demonstrates the best performance across three tasks: STS, retrieval, and summary retrieval, compared to  other models. Additionally, the BGE-m3 model achieves the highest accuracy in the rerank task and exhibits an accuracy close to that of the Jina model in the retrieval task. Therefore, we identify this model as a tailored model for the  information retrieval task.
Moreover, in the classification, clustering, and Pair classification tasks, we observe that the sentence-transformer-parsbert, TookaBERT, and Tooka-SBERT models, which are trained specifically for the Persian language, achieve the highest accuracies, respectively.

\begin{table*}[h!]
\centering
\begin{adjustbox}{width=\textwidth}
\begin{tabular}{c|cc|ccccccc}

\textbf{} & Size (M) & Avg. & Class. & Cluster. & PairClass. & Rerank. & Retriv. & STS & SumRet. \\
\hline
        
        sentence-transformer-parsbert-fa & 163 & 37.93 & 53.06 & \textbf{64.83} & 70.55 & 39.9 & 8.98 & 55.07 & 14.71  \\
        RoBERTa-WLNI & 110 & 37.97 & 54.87 & 58.61 & 71.1 & 44.74 & 9.91 & 54.92 & 5.71  \\
        BERT-WLNI & 110 & 38.15 & 54.59 & 60.32 & 71.07 & 45.72 & 9.82 & 56.27 & 6.35  \\
        faBert & 124 & 40.62 & 60.82 & 55.18 & 68.73 & 50.58 & 13.92 & 52.01 & 9.32  \\ 
        ParsBERT & 110 & 40.63 & 65.13 & 56.15 & 69.14 & 46.38 & 9.91 & 60.97 & 7.09  \\
        paraphrase-multilingual-MiniLM-L12-v2 & 118 & 46.62 & 55.58 & 58.12 & 79.9 & 55.82 & 23.08 & 67.24 & 31.91  \\
        LaBSE & 471 & 48.47 & 62.02 & 56.56 & 78.88 & 55.74 & 21.2 & 73.06 & 47.51  \\
        TookaBERT-Base & 123 & 41.17 & \textbf{65.54} & 55.72 & 70.69 & 44.18 & 10.51 & 61.89 & 8.29  \\
        Tooka-SBERT & 353 & 52.74 & 61.29 & 56.45 & \textbf{87.04} & 58.29 & 27.86 & 76.41 & 59.06  \\
        GTE-multilingual-base & 305 & 57.14 & 58.6 & 57.28 & 84.57 & 69.72 & 41.22 & 75.75 & 60.88  \\
        multilingual-e5-base & 278 & 57.03 & 59.97 & 56.52 & 84.04 & 72.07 & 41.2 & 74.45 & 54.58  \\
        multilingual-e5-large & 560 & 58.44 & 61.7 & 57.19 & 84.04 & 74.34 & 42.98 & 75.38 & 56.61  \\
        BGE-m3-unsupervised & 568 & 56.6 & 60.99 & 59.62 & 83.07 & 69.74 & 38.14 & 74.47 & 61.67  \\
        BGE-m3 & 567 & 59.1 & 61.74 & 57.73 & 85.21 & \textbf{74.56} & 43.38 & 76.35 & 61.07  \\
        Jina-embeddings-v3 & 572 & \textbf{59.28} & 62.97 & 59.15 & 83.71 & 61.26 & \textbf{43.51} & \textbf{78.65} & \textbf{65.5}  \\ 

\end{tabular}
\end{adjustbox}
\caption{Evaluation of various text embedding models on the FaMTEB benchmark. The numbers indicate the percentage of success rate.}
\label{tab:faMTEB_results}
\end{table*}

\section{Conclusion}
In this study, we presented FaMTEB, a comprehensive benchmark for evaluating Persian text embeddings, building upon the massive text embedding benchmark (MTEB). Our contribution spans several dimensions, including the introduction of $63$ datasets across seven diverse tasks, the addition of a novel task (summary retrieval), and the creation of numerous new Persian-language NLP datasets for training and evaluation. By integrating datasets related to chatbot challenges and RAG systems into MTEB for the first time, we provide a tailored evaluation framework that aligns with emerging applications of text embedding models. The performance evaluation of Persian and multilingual embedding models further underscores the benchmark's utility. This open-source benchmark, encompassing datasets, code and a public leaderboard, establishes a robust foundation for advancing NLP research in the Persian language. We hope FaMTEB inspires further innovation and enhances the development of more effective Persian-language models.

\section{Limitations}
Despite advancements in Persian natural language processing (NLP), the availability of human-annotated datasets remains limited. The scarcity of such data poses a significant challenge in training and evaluating high-quality models across various NLP tasks. Addressing this limitation requires extensive efforts to construct diverse and representative datasets tailored to different linguistic problems. This challenge is particularly evident in tasks such as semantic textual similarity, reranking, and summarization, where high-quality labeled data is essential for achieving reliable performance.

Due to the high costs associated with evaluating text embedding models that provide their APIs on a paid basis, such as \textit{text-embedding-3-large}, as well as the resource-intensive nature of serving and evaluating some other open-source text embedding models, we have not yet included certain models in the leaderboard. We are gradually adding these models over time.

\bibliography{refs}

\appendix

\section{Dataset}
In this section, we review the complete set of datasets used in the FaMTEB benchmark. Table \ref{tab:faMTEB_count} provides the number of data for each dataset.

\begin{figure*}[t]
\includegraphics[width=\textwidth]{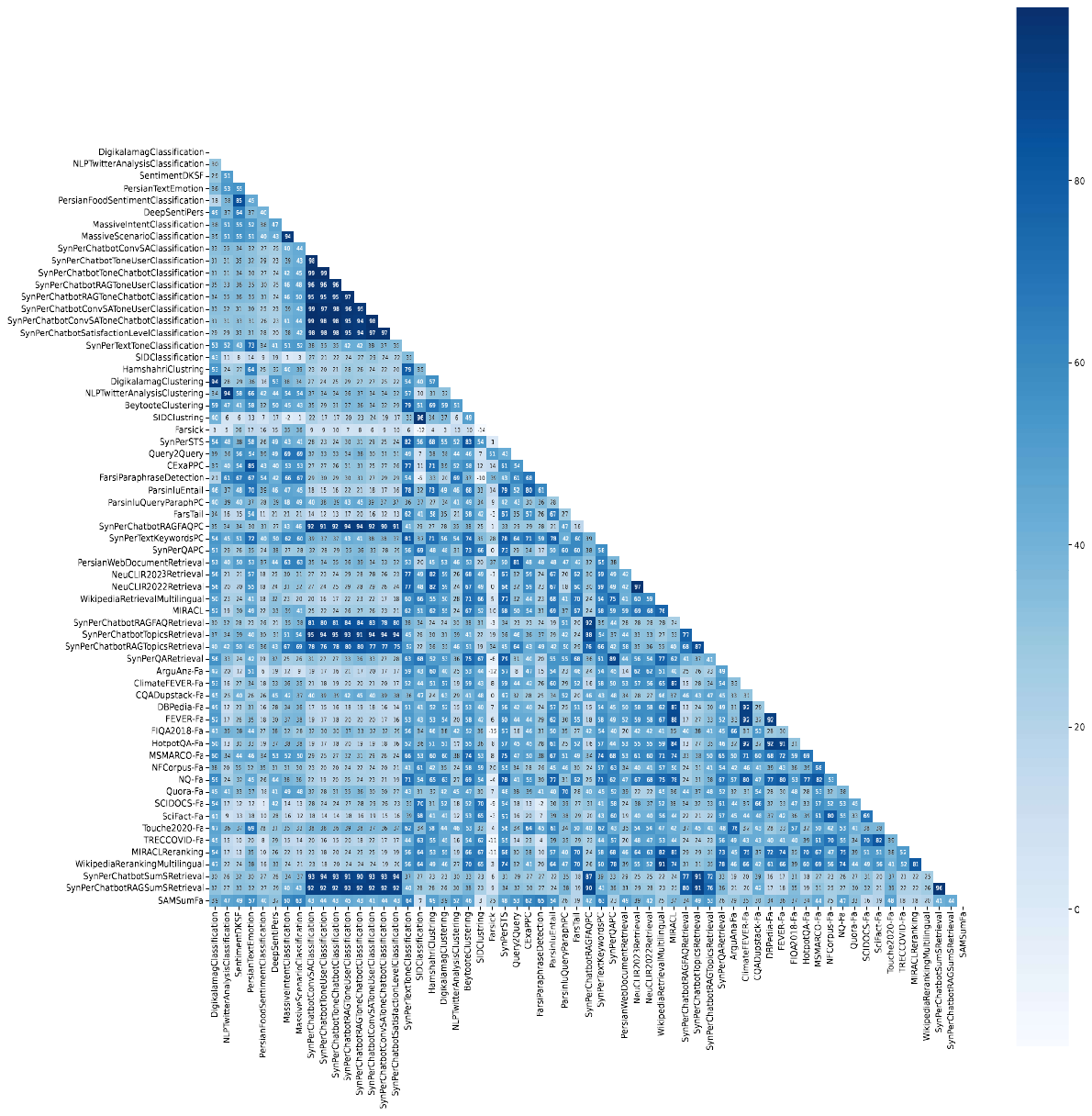}
\centering
\caption{Similarity chart of different MTEB datasets.}
\label{fig:dataset_correlation}
\end{figure*}

Figure \ref{fig:dataset_correlation} presents the similarity chart of different datasets. To construct this chart, we first randomly selected 100 samples from each dataset and computed the average vector of these samples. Next, we calculated the cosine similarity between the resulting vectors for the datasets, multiplied the values by 100, and rounded them to obtain the final chart.

For vector representation, we utilized the Jina-embeddings-v3 model, which demonstrated the highest performance among the evaluated models in both the STS task and overall average performance (see Table \ref{tab:faMTEB_results}).

As observed, the FaMTEB datasets exhibit a wide range of similarity values. In general, most similarity scores between datasets are relatively low, indicating the diverse nature of the FaMTEB datasets.

\begin{figure*}[t]
\includegraphics[width=\textwidth]{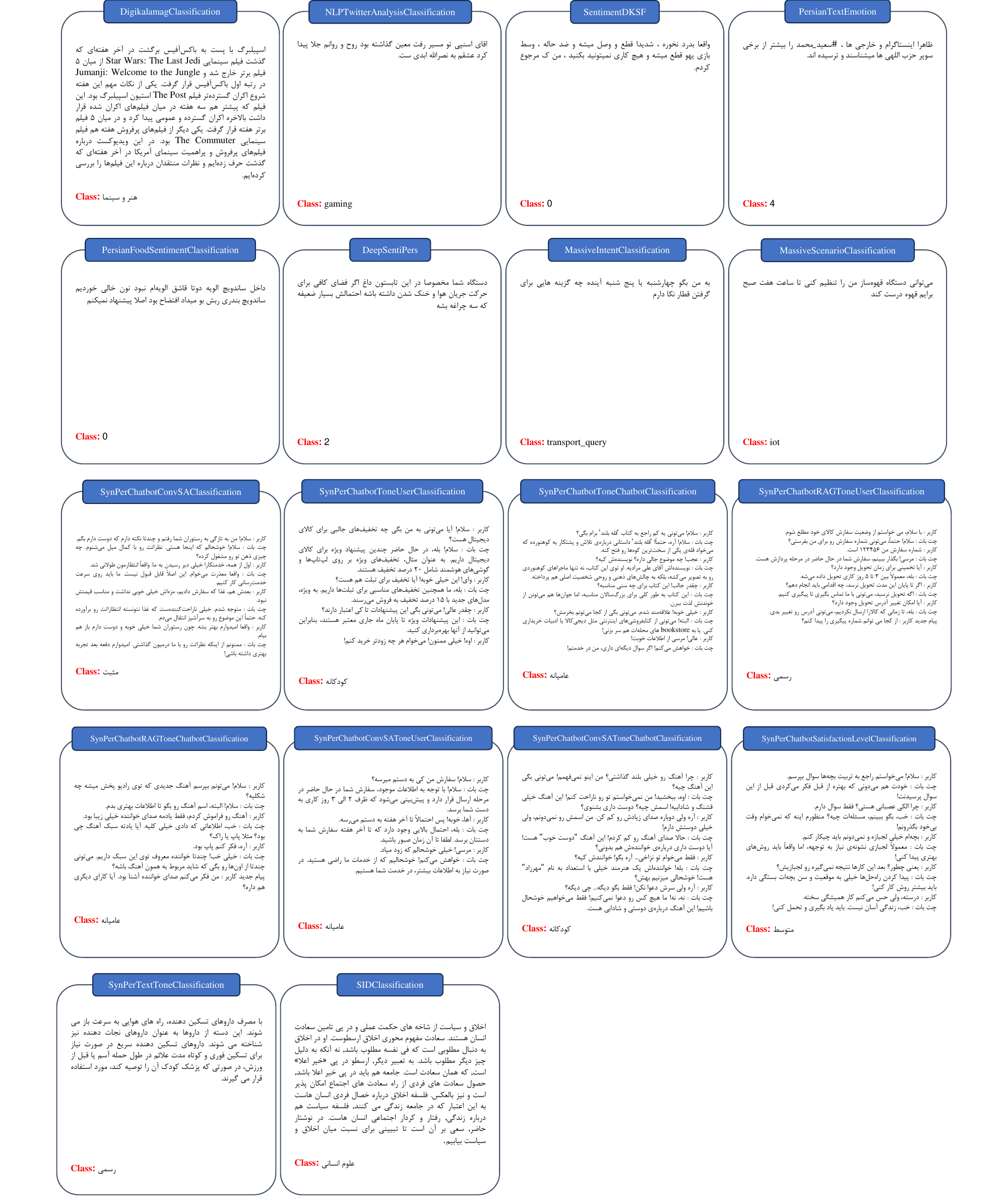}
\centering
\caption{Example of classification datasets.}
\label{fig:ClassificationExample}
\end{figure*}

\begin{figure*}[t]
\includegraphics[width=\textwidth]{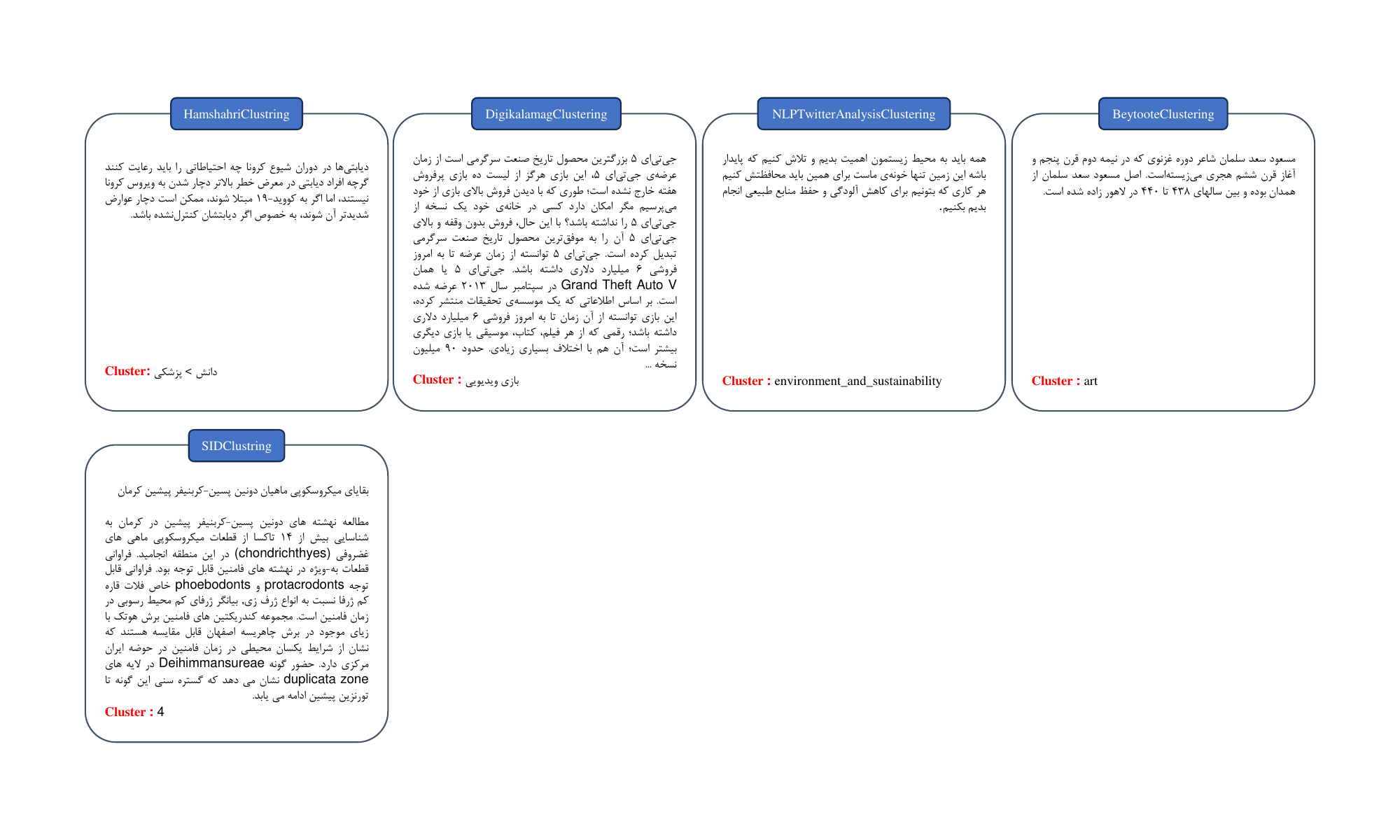}
\centering
\caption{Example of clustering datasets.}
\label{fig:ClusteringExample}
\end{figure*}

\begin{figure*}[t]
\includegraphics[width=\textwidth]{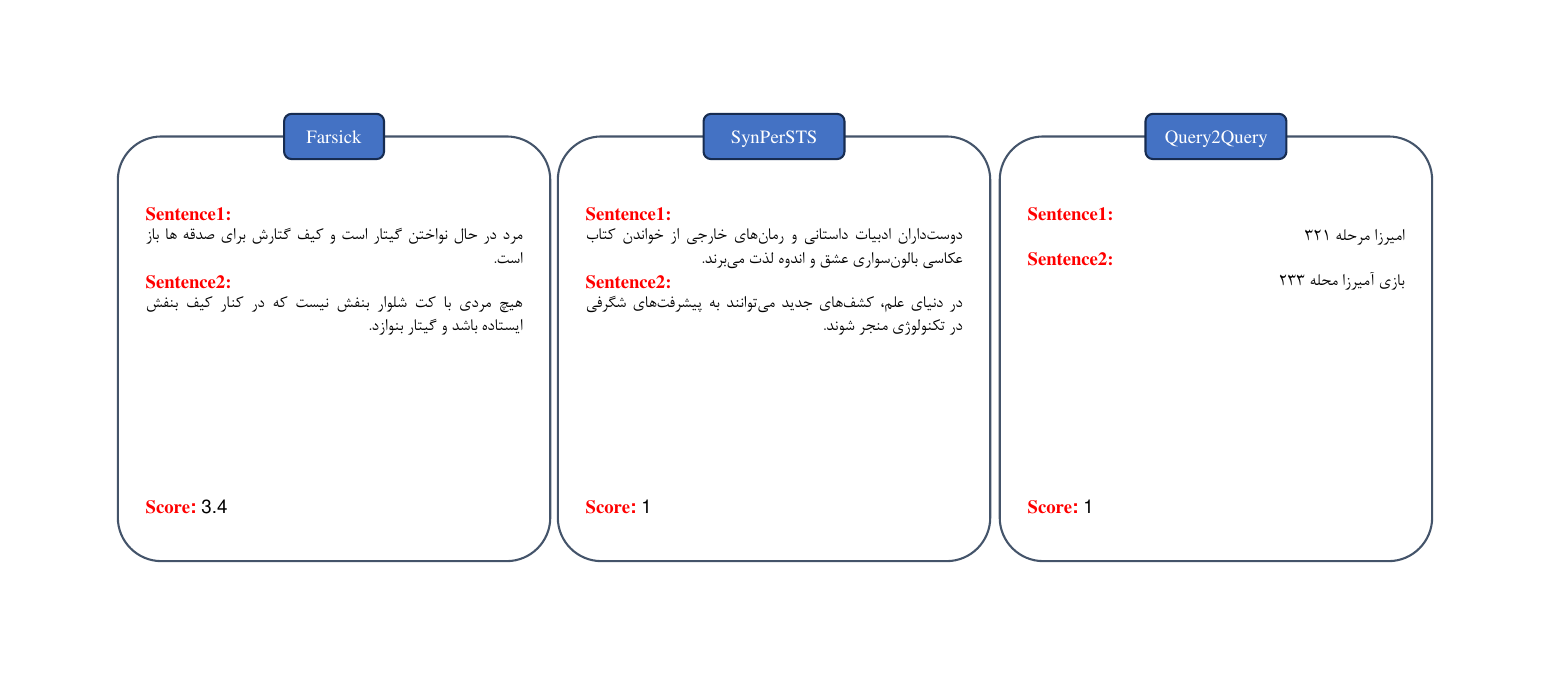}
\centering
\caption{Example of STS datasets.}
\label{fig:STSExample}
\end{figure*}

\begin{figure*}[t]
\includegraphics[width=\textwidth]{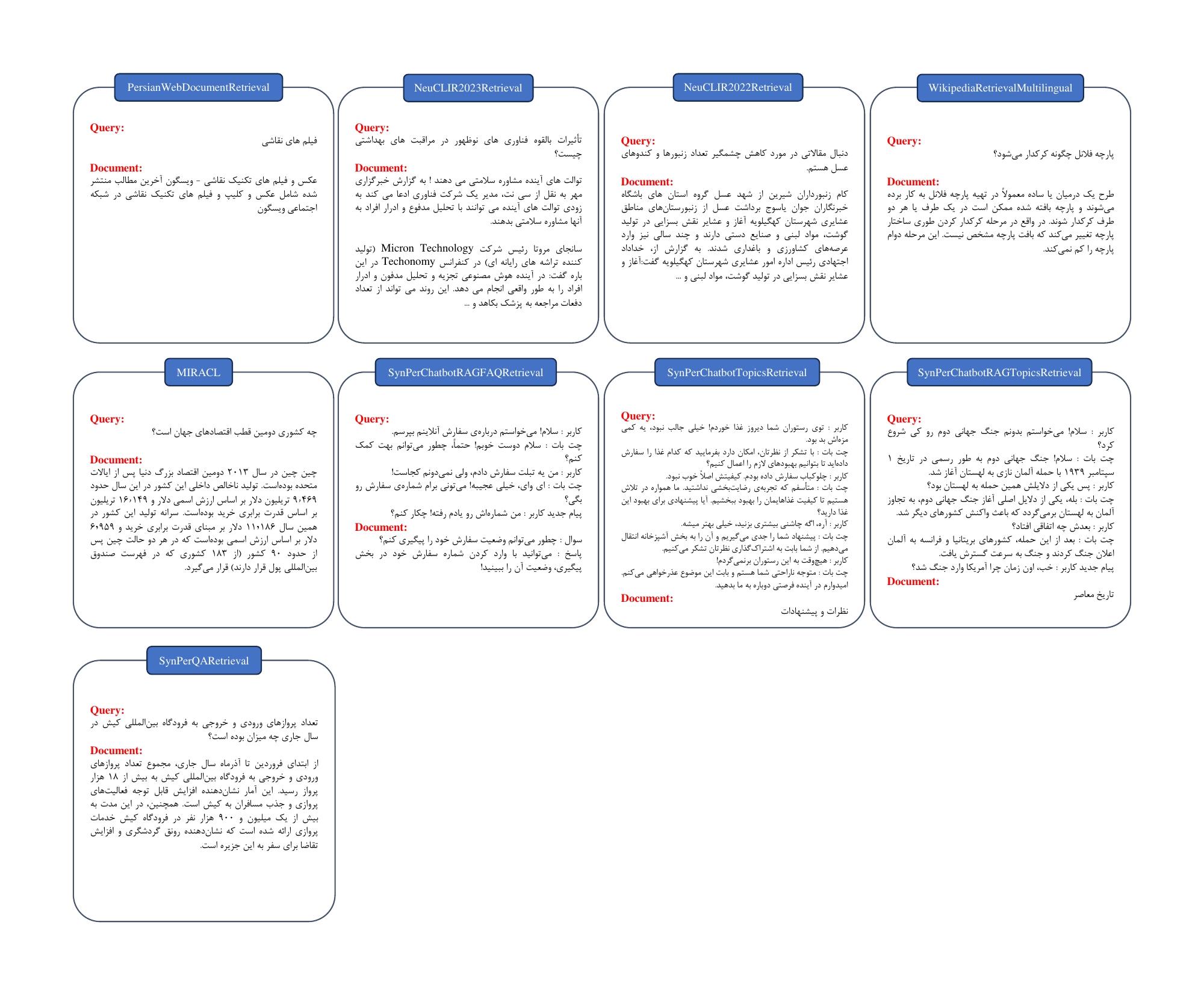}
\centering
\caption{Example of retrieval datasets.}
\label{fig:RetrievalExample}
\end{figure*}

\begin{figure*}[t]
\includegraphics[width=\textwidth]{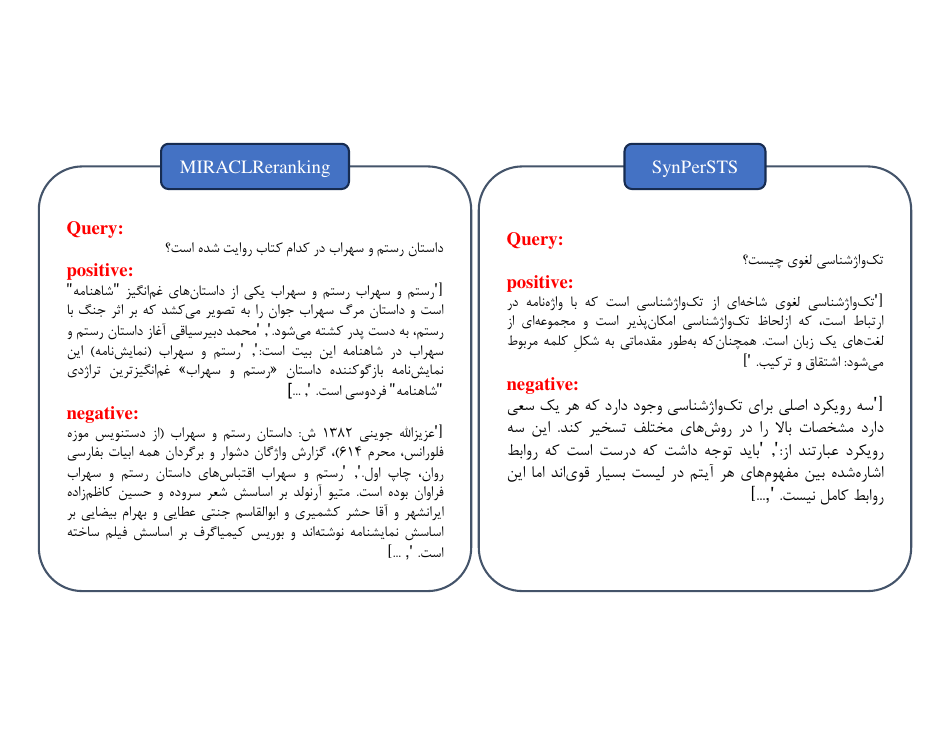}
\centering
\caption{Example of reranking datasets.}
\label{fig:RerankingExample}
\end{figure*}

\begin{figure*}[t]
\includegraphics[width=\textwidth]{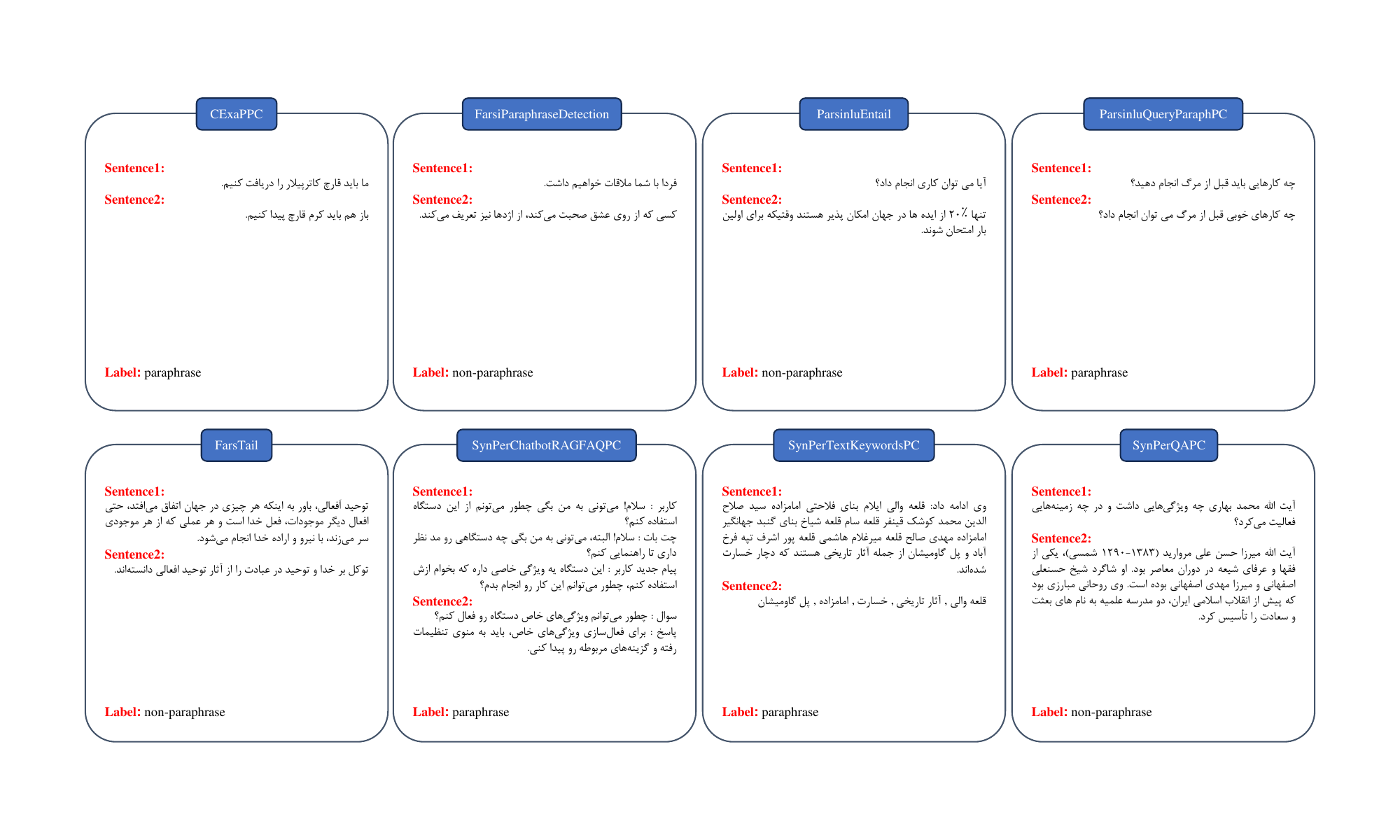}
\centering
\caption{Example of pair classification datasets.}
\label{fig:PCExample}
\end{figure*}

\begin{figure*}[t]
\includegraphics[width=\textwidth]{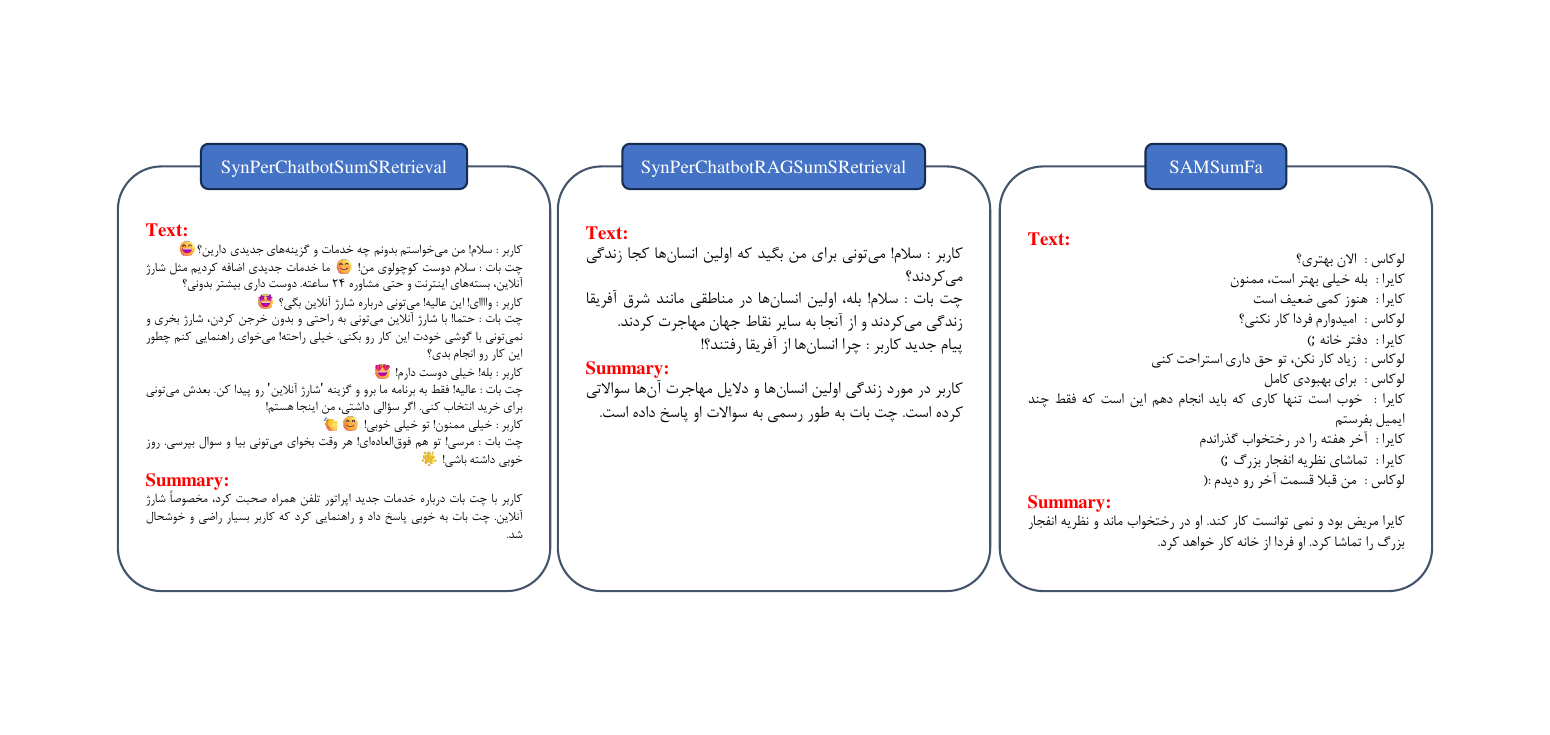}
\centering
\caption{Example of summary retrieval datasets excluding BEIR-Fa datasets.}
\label{fig:SumRetExample}
\end{figure*}

We have also provided a sample from each of the different datasets available in this benchmark in Figures \ref{fig:ClassificationExample}, \ref{fig:ClusteringExample}, \ref{fig:STSExample}, \ref{fig:RetrievalExample}, \ref{fig:RerankingExample}, \ref{fig:PCExample}, and \ref{fig:SumRetExample}.

\begin{figure}[ht]
\includegraphics[width=\columnwidth]{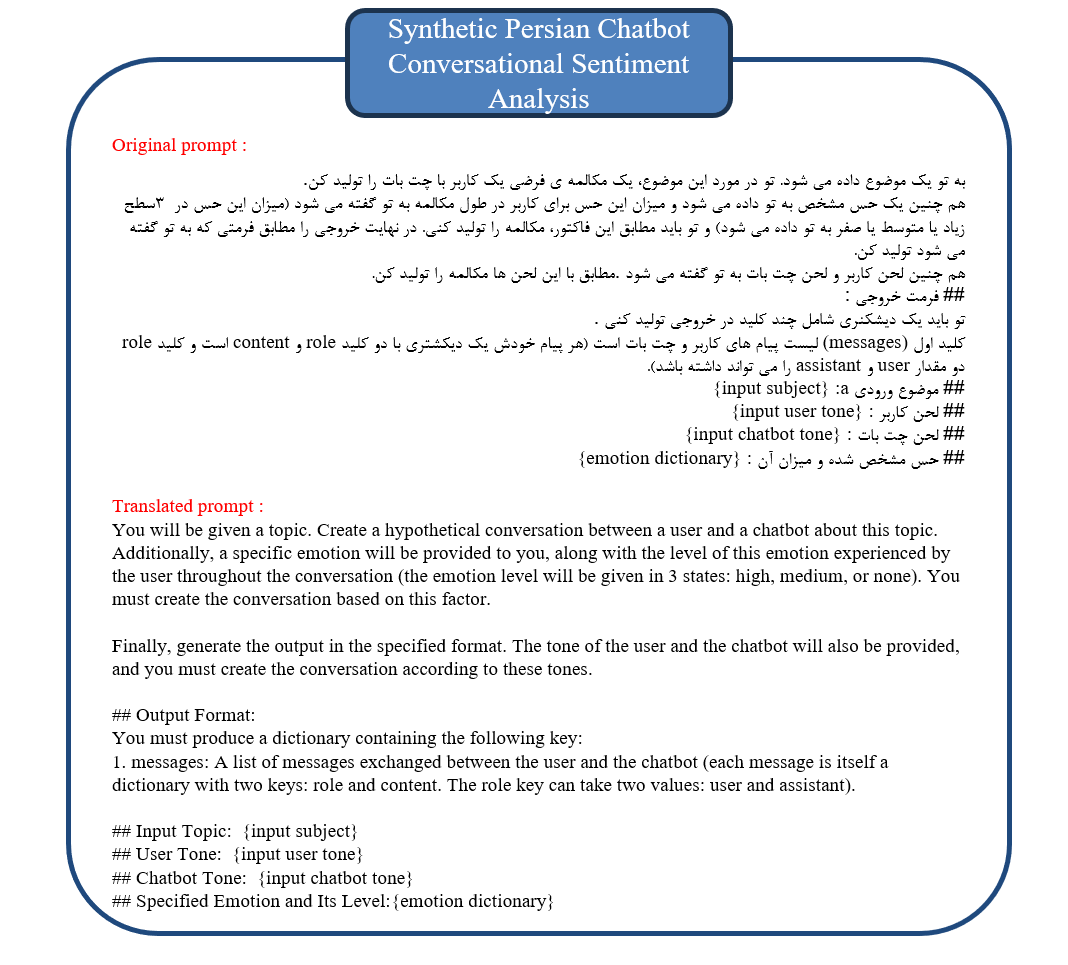}
\centering
\caption{This figure illustrates the prompts used to generate the Synthetic Persian Chatbot Conversational Sentiment Analysis dataset. This dataset receives the chat subject, user tone, chatbot tone, and user emotion as input, corresponding to the placeholders "input subject", "input user tone", "input chatbot tone", and "emotion dictionary", respectively.}
\label{fig:SynPerChatbotConvSAToneClassification}
\end{figure}

\begin{figure*}[t]
\includegraphics[width=\textwidth]{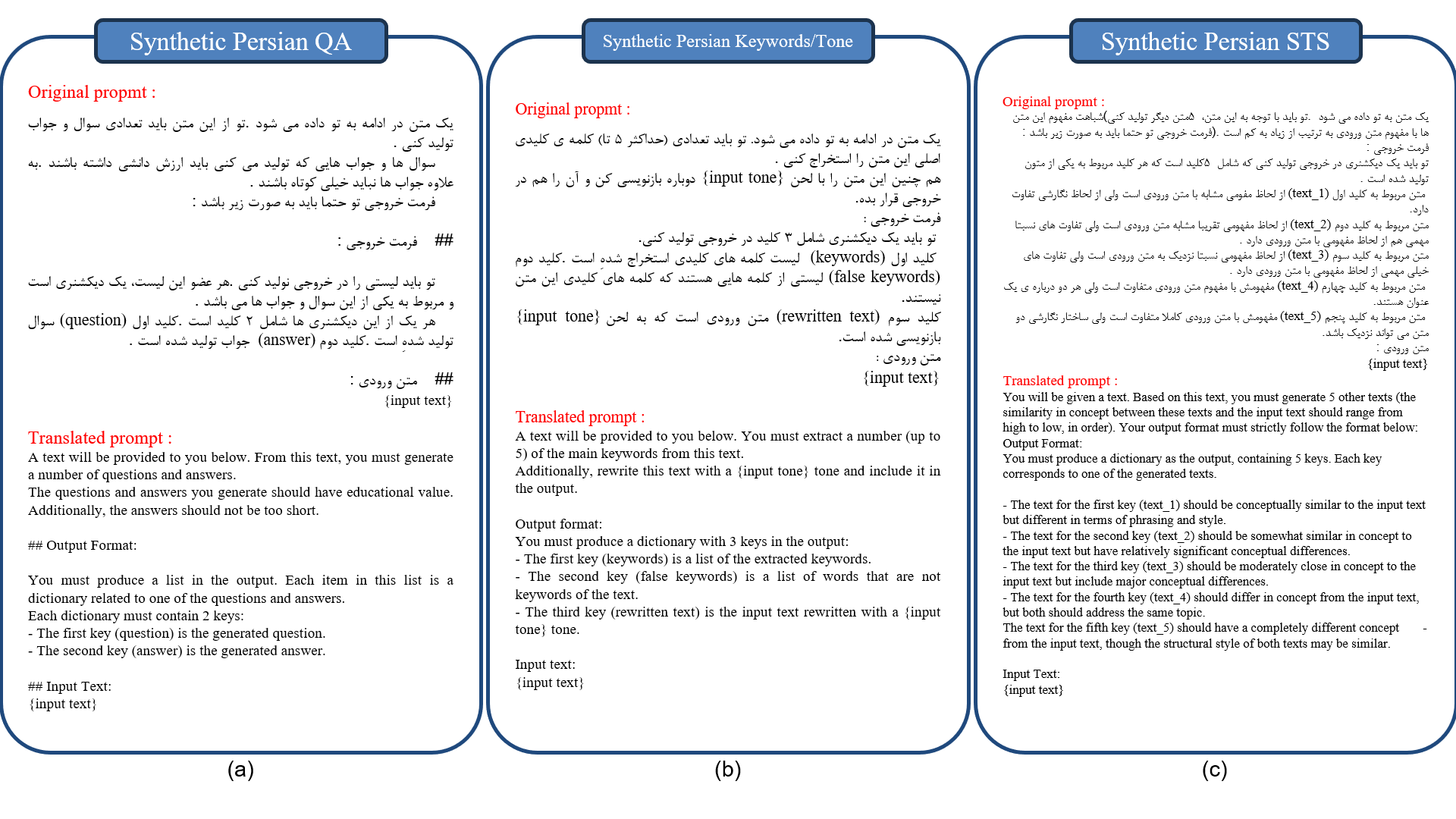}
\centering
\caption{This figure illustrates the prompts used to generate the three datasets: a) Synthetic Persian QA, b) Synthetic Persian Keywords/Tone, and c) Synthetic Persian STS. The prompts are in Persian, with their translations included below each prompt. In the prompts, the placeholder "input text" is replaced with the text intended to generate the data. Additionally, in prompt b, there is an "input tone" placeholder to specify the desired tone.}
\label{fig:SynPer}
\end{figure*}

\begin{figure*}[t]
\includegraphics[width=\textwidth]{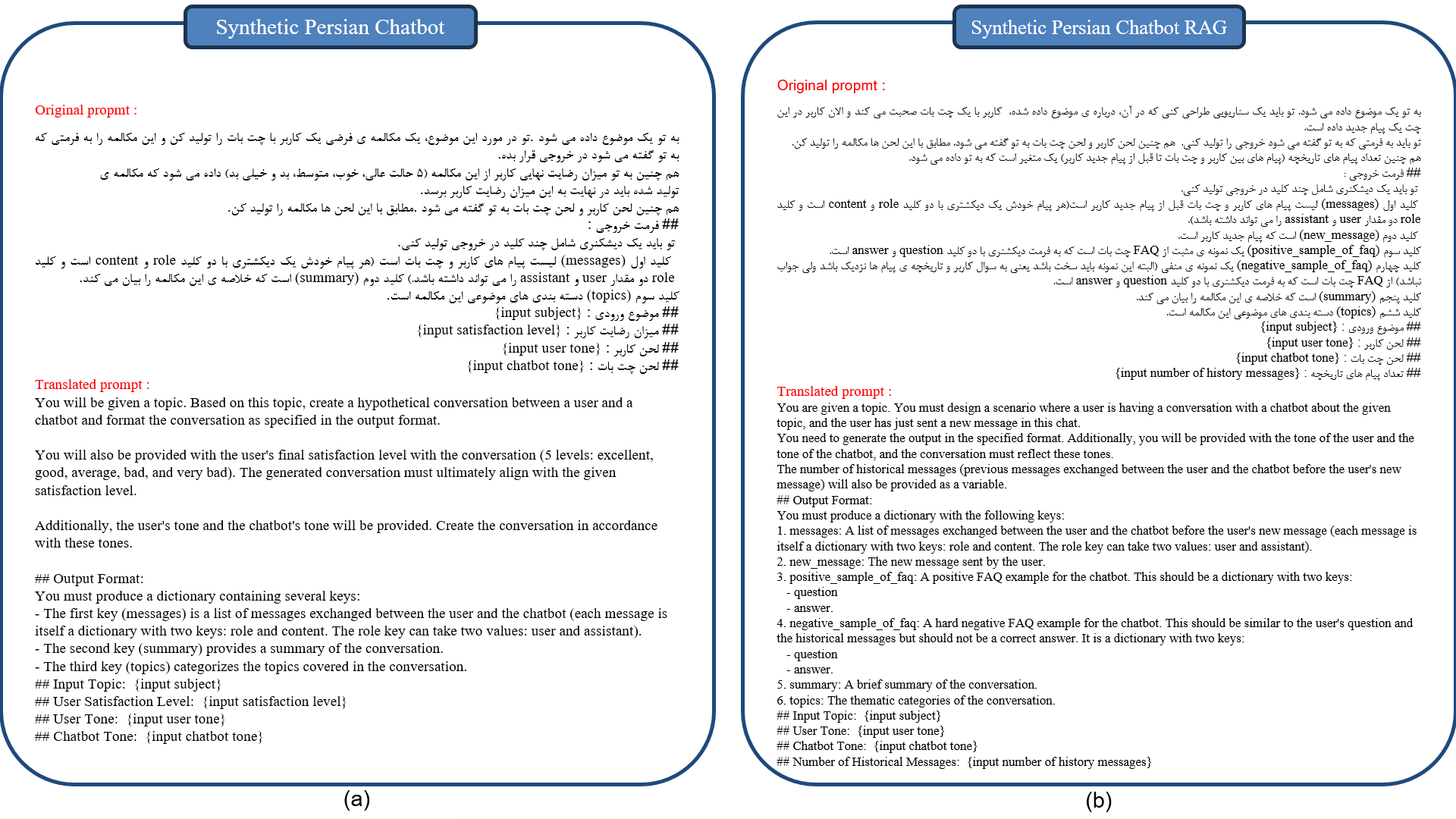}
\centering
\caption{This figure illustrates the prompts used to generate the two datasets: a) Synthetic Persian Chatbot and b) Synthetic Persian Chatbot RAG. Both datasets receive the chat subject, user tone, and chatbot tone as input, corresponding to the placeholders "input subject", "input user tone", and "input chatbot tone", respectively. Additionally, in prompt a, the user's satisfaction level is provided through the placeholder "input satisfaction level", and in prompt b, the number of messages included in the chat history is specified using the placeholder "input number of history messages".}
\label{fig:SynPerChatbot}
\end{figure*}

The distribution of sources used to construct the \textit{synthetic Persian STS}, \textit{synthetic Persian keywords}, and \textit{synthetic Persian tone} datasets is detailed in Table \ref{tab:doc_num}. Furthermore, the set of prompts employed for generating the synthetic datasets is depicted in Figures \ref{fig:SynPer}, \ref{fig:SynPerChatbot}, and \ref{fig:SynPerChatbotConvSAToneClassification}.

\label{DatasetsDetails}

\subsection{Classification}
\label{Classification}
\textbf{DigikalamagClassification~\citep{Farahani2020ParsBERTTM}} The Digikala Magazine dataset (DigikalamagClassification) consists of 8,515 articles scraped from the Digikala Online Magazine. The dataset is organized into seven distinct classes: Video Games, Shopping Guide, Health \& Beauty, Science \& Technology, General, Art \& Cinema, and Books \& Literature.

\textbf{NLPTwitterAnalysisClassification~\citep{nlp_twitter_analysis}} The NLPTwitterAnalysisClassification is a classification dataset that categorizes various tweets into 26 distinct classes based on the topics they discuss.
 
\textbf{SentimentDKSF~\citep{hezar2023}} This dataset is composed of comments collected from the Digikala~\footnote{digikala.com} and SnappFood~\footnote{snappfood.ir} websites. Each comment is annotated with one of three labels: Positive, Negative, or Neutral, indicating the user's level of satisfaction.

\textbf{PersianTextEmotion~\citep{persian_text_emotion}}
PersianTextEmotion is a Persian sentiment analysis dataset containing 6,948 sentences, each annotated with one of six distinct emotions: joy, sadness, anger, disgust, fear, or surprise.

\textbf{PersianFoodSentimentClassification~\citep{Farahani2020ParsBERTTM}} The PersianFoodSentimentClassification dataset consists of 70,000 user comments collected from SnappFood, an online food delivery platform. The dataset is designed for polarity classification and includes two sentiment labels: (0) Happy and (1) Sad. This dataset provides a comprehensive resource for sentiment analysis, particularly in assessing customer satisfaction and experience within the context of online food delivery services.

\textbf{DeepSentiPers~\citep{PourmostafaRoshanSharami2020DeepSentiPersND}} The DeepSentiPers dataset is an enhanced and balanced version of the SentiPers dataset~\citep{Hosseini2018SentiPersAS}, containing 12,138 user opinions on digital products. It is annotated with five sentiment classes: two positive (Happy and Delighted), two negative (Furious and Angry), and one Neutral class. This dataset can be utilized for both multi-class and binary sentiment classification tasks. For binary classification, the Neutral class is excluded, focusing the analysis on positive and negative sentiments.

\textbf{MassiveIntentClassification~\citep{fitzgerald-etal-2023-massive}} 
This dataset comprises a collection of Amazon Alexa virtual assistant utterances, each annotated with its corresponding intent. One of 60 possible intents is assigned as the label for each user utterance. The dataset is multilingual, supporting 51 languages, including Persian.

\textbf{MassiveScenarioClassification~\citep{fitzgerald-etal-2023-massive}}
This dataset comprises a collection of Amazon Alexa virtual assistant utterances, each annotated with its corresponding scenario. One of 60 possible intents is assigned as the label for each user utterance. The dataset is multilingual, supporting 51 languages, including Persian.

\textbf{SynPerChatbotConvSAClassification (Some Emotion)} For each of the nine emotions in the \textit{Synthetic Persian Chatbot Conversational Sentiment Analysis}, a separate classification dataset was created.  
The samples for each dataset were selected from the \textit{Synthetic Persian Chatbot Conversational Sentiment Analysis Dataset}, focusing on conversations that exhibit the specific emotion of interest. Each sample represents a conversation between the user and the chatbot, and is labeled with one of two classes:
\begin{itemize}
    \item \textbf{Negative}: When the intensity level of the emotion is zero.
    \item \textbf{Positive}: When the intensity level of the emotion is moderate or high.
\end{itemize}

\textbf{SynPerChatbotToneUserClassification}
This is a classification dataset derived from the \textit{Synthetic Persian Chatbot}, which classifies the user's tone in the conversation between the user and the chatbot.

\textbf{SynPerChatbotToneChatbotClassification}
This is a classification dataset derived from the \textit{Synthetic Persian Chatbot}, which classifies the chatbot's tone in the conversation between the user and the chatbot.

\textbf{SynPerChatbotRAGToneUserClassification}
This is a classification dataset derived from the \textit{Synthetic Persian Chatbot RAG}, which classifies the user's tone in the conversation between the user and the chatbot.

\textbf{SynPerChatbotRAGToneChatbotClassification}
This is a classification dataset derived from the \textit{Synthetic Persian Chatbot RAG}, which classifies the chatbot's tone in the conversation between the user and the chatbot.

\textbf{SynPerChatbotConvSAToneUserClassification}
This is a classification dataset derived from the \textit{Synthetic Persian Chatbot Conversational Sentiment Analysis}, which classifies the user's tone in the conversation between the user and the chatbot.

\begin{adjustbox}{width=\columnwidth}
\textbf{SynPerChatbotConvSAToneChatbotClassification}
\end{adjustbox}
This is a classification dataset derived from the \textit{Synthetic Persian Chatbot Conversational Sentiment Analysis}, which classifies the chatbot's tone in the conversation between the user and the chatbot.

\textbf{SynPerChatbotSatisfactionLevelClassification}
This is a classification dataset derived from the \textit{Synthetic Persian Chatbot Dataset}, which classifies the user's satisfaction level in the conversation between the user and the chatbot.

\textbf{SynPerTextToneClassification}
This is a classification dataset derived from the \textit{Synthetic Persian Tone Dataset}. The samples consist of rewritten texts, and the classes correspond to the tones of the rewritten texts.

\textbf{SIDClassification}
As detailed in Section \ref{SID}, the SIDClassification dataset is derived from the SID\footnote{https://www.sid.com/} website. In this dataset, the titles and abstracts of articles are used as input texts, which are classified according to the website's predefined category labels, with these categories serving as the class labels.

\subsection{Clustering}
\textbf{BeytooteClustering} As explained in Section \ref{BeytooteClustering}, BeytooteClustering is a clustering dataset collected from the Beytoote\footnote{https://www.beytoote.com/} website. The BeytooteClustering dataset consists of 19 categories, as detailed in Table \ref{tab:beytoote}.

\textbf{DigikalamagClustering} This dataset is the same as the DigikalamagClassification dataset described in Section \ref{Classification}, with the distinction that it is used here for the clustering task.

\textbf{HamshahriClustring~\citep{farsi_news}} The Hamshahri dataset is a subset of the Farsi-news dataset, extracted from the RSS feeds of two prominent Farsi news agency websites: Hamshahri and RadioFarda. Each entry in this dataset consists of the title, summary, URL, and tags of the respective news page. For the Hamshahri dataset specifically, the title and summary are concatenated to form the input text, while the tag associated with each page is used as the classification label.

\textbf{NLPTwitterAnalysisClustering} This dataset is the same as the NLPTwitterAnalysisClassification dataset described in Section \ref{Classification}, with the distinction that it is used here for the clustering task.

\textbf{SIDClustring} As described in Section \ref{SID}, SIDClustring is a clustering dataset sourced from the SID\footnote{https://www.sid.com/} website. In this dataset, the title and abstract of articles serve as input texts, which are categorized based on the website's predefined category labels.

\subsection{STS}
\textbf{Farsick~\citep{9721521}}
FarSick is a Semantic Textual Similarity (STS) dataset designed for the Persian language. The dataset comprises approximately 10,000 sentence pairs, each meticulously annotated for semantic relatedness, meaning alignment, and entailment relations between the sentence pairs. FarSick was developed by translating and adapting the sentence pairs from the original SICK dataset, ensuring relevance and applicability to the Persian linguistic context.

\textbf{SynPerSTS}
As described in \ref{SynPerSTS}, we utilize GPT-4o-mini to generate synthetic data for the Semantic Textual Similarity (STS) task.

\begin{figure}[ht]
    \centering
    \begin{subfigure}[b]{\linewidth}
        \centering
        \includegraphics[width=\textwidth]{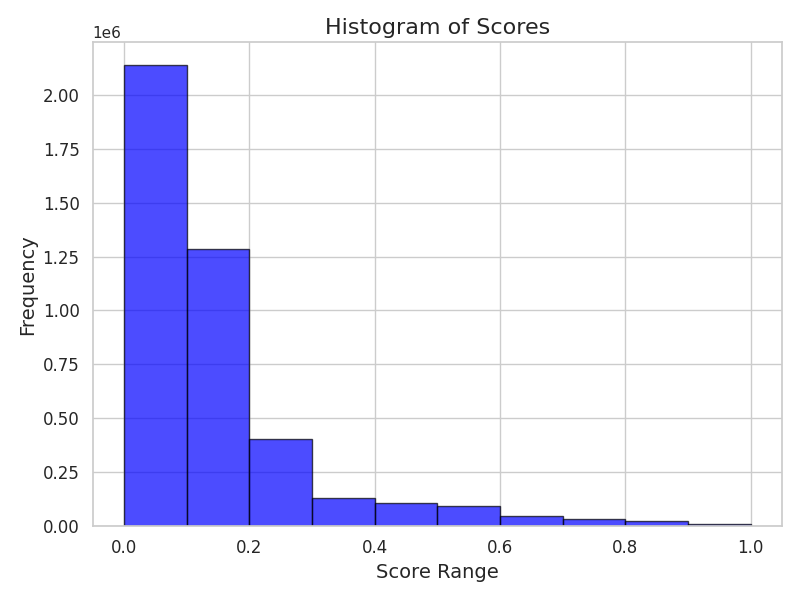}  
        \caption{}
        \label{fig:subfig1}
    \end{subfigure}
    
    \vspace{0.5cm}  
    
    \begin{subfigure}[b]{\linewidth}
        \centering
        \includegraphics[width=\textwidth]{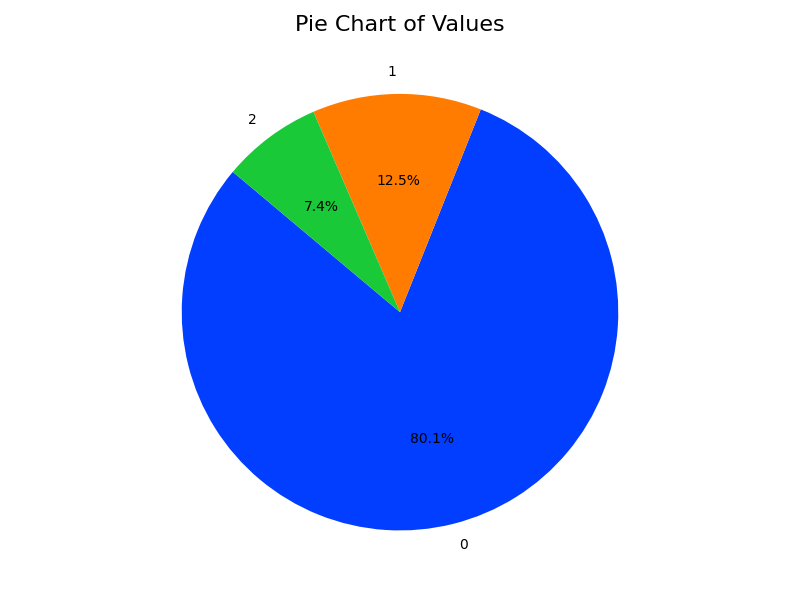}  
        \caption{}
        \label{fig:subfig2}
    \end{subfigure}
    
    \caption{Figure a shows the distribution of the Query to Query data based on similarity scores. Figure b illustrates the distribution of the Query to Query data according to the labels assigned, ranging from 0 to 2.}
    \label{fig:q2q}
\end{figure}

\textbf{Query2Query}
As detailed in Section \ref{Query2Query}, this is a Semantic Textual Similarity (STS) dataset comprising search engine queries, categorized into three levels of similarity. The distribution of labels is shown in Figure \ref{fig:q2q}.

\subsection{Summary Retrieval}
\textbf{SynPerChatbotSumSRetrieval}
This dataset is derived from the \textit{Synthetic Persian Chatbot Dataset}.  
It is designed to evaluate the model's ability to retrieve summaries of conversations between the user and the chatbot.

\textbf{SynPerChatbotRAGSumSRetrieval}
This dataset is derived from the \textit{Synthetic Persian Chatbot RAG Dataset}.  
It is designed to evaluate the model's ability to retrieve summaries of user-chatbot conversations, including conversations that may not have been completed.

\textbf{SAMSumFa} 
The SAMSum dataset is a dataset designed for abstractive dialogue summarization. It is composed of real-world-like chat conversations along with human-written summaries. SAMSumFa is a translation of the SAMSum dataset using the exact method proposed for BEIR in Section \ref{BEIR}.

\subsection{Retrieval}
\textbf{ArguAna-Fa}  
ArguAna-Fa retrieves the best counterargument to an argument. This dataset aids in assessing the relevance and quality of counterarguments provided in debates or discussions.

\textbf{ClimateFEVER-Fa}  
ClimateFEVER-Fa verifies climate claims using the FEVER Wiki corpus, with claims as queries and evidence retrieval. The dataset is used to evaluate the effectiveness of systems in fact-checking climate-related statements based on available evidence.

\textbf{CQADupstack-Fa}  
CQADupstack-Fa is a community question-answering dataset with queries from 12 StackExchange subforums, evaluated using retrieval of duplicate queries, reporting mean scores in BEIR. It focuses on identifying and retrieving relevant information based on user questions across a variety of topics.

\textbf{DBPedia-Fa}  
DBPedia-Fa is an entity retrieval dataset with heterogeneous queries, focusing on retrieving entities from the English DBpedia corpus. It is valuable for tasks that involve querying knowledge bases and retrieving structured data based on user inputs.

\textbf{FEVER-Fa}  
FEVER-Fa dataset supports automatic fact-checking by retrieving evidence from pre-processed Wikipedia abstracts using original paper splits as queries. This dataset is particularly useful for evaluating models that perform fact-checking tasks, ensuring they can retrieve reliable sources of evidence.

\textbf{FiQA2018-Fa}  
FiQA2018-Fa focuses on opinion-based QA, using financial data from StackExchange Investment posts. The dataset evaluates the ability of systems to answer subjective questions related to financial markets, opinions, and trends.

\textbf{HotpotQA-Fa}  
HotpotQA-Fa requires obtaining the correct answer to a question by reasoning in multiple paragraphs. It challenges models to combine information from several documents to arrive at accurate answers, thus testing systems' multi-hop reasoning abilities.

\textbf{MSMARCO-Fa}  
MSMARCO-Fa is an information retrieval dataset comprising queries from Bing logs and various web documents, where each query is associated with both relevant and irrelevant documents. This dataset is used to evaluate information retrieval systems, particularly their ability to rank documents by relevance to the query.

\textbf{NFCorpus-Fa}  
NFCorpus-Fa includes queries from NutritionFacts and an annotated medical corpus from PubMed. The dataset supports research in information retrieval in the medical domain, where accurate and relevant information is critical for health-related queries.

\textbf{NQ-Fa}  
NQ-Fa consists of a set of Google search engine queries and their corresponding answers from Wikipedia, which have been human-annotated. This dataset is designed for training and evaluating models on natural question answering tasks.

\textbf{Quora-Fa}  
Quora-Fa consists of the Quora corpus, where duplicate queries have been identified. It is primarily used to assess the ability of systems to detect duplicate questions and provide consistent answers.

\textbf{SCIDOCS-Fa}  
SCIDOCS-Fa is a citation prediction task aimed at identifying directly cited scientific articles based solely on the title of the citing article. The dataset facilitates research into automatic citation prediction and understanding of scientific knowledge connections.

\textbf{SciFact-Fa}  
SciFact-Fa dataset is a scientific fact verification benchmark that evaluates the ability of models to verify claims using evidence from the scientific literature, focusing on claim-evidence alignment and factual correctness. It is critical for fact-checking tasks in the scientific domain.

\textbf{Touche2020-Fa}  
Touche2020-Fa is a dataset designed for argument retrieval tasks. The goal is to identify and retrieve relevant arguments to counter or support claims based on a given premise. It evaluates models' ability to engage in argumentative reasoning by considering both the quality and relevance of retrieved arguments.

\textbf{TRECCOVID-Fa}  
TRECCOVID-Fa is an ad-hoc search challenge using the July 16, 2020 CORD-19 dataset with cumulative judgments and query descriptions from the original task. This dataset is used for evaluating information retrieval systems on COVID-19-related content.

\noindent \textbf{Note:} All of these datasets have been translated from the original BEIR datasets to the Persian language as explained in Section \ref{BEIR}.

\textbf{SynPerChatbotRAGFAQRetrieval}
This dataset is derived from the \textit{Synthetic Persian Chatbot RAG Dataset}. It was created to evaluate the model's retrieval capabilities in a RAG-based chatbot system. By incorporating elements such as message history, the new user message (which may or may not be a follow-up), and a question-answer format for the documents, it simulates a real-world retrieval scenario in a RAG chatbot system.

\textbf{SynPerChatbotTopicsRetrieval}
This dataset is derived from the \textit{Synthetic Persian Chatbot Dataset}. It was created to evaluate the model's retrieval capabilities for identifying the topics of conversations between the user and the chatbot.

\textbf{SynPerChatbotRAGTopicsRetrieval}
This dataset is derived from the \textit{Synthetic Persian Chatbot RAG Dataset}. It was created to evaluate the model's retrieval capabilities for identifying the topics of user-chatbot conversations, including conversations that have not necessarily been completed.

\textbf{SynPerQARetrieval}
This dataset is derived from the \textit{Synthetic Persian QA Dataset}.  
It is designed to evaluate the model's retrieval capabilities when the query is a question, and the documents are in the format of answers.

\textbf{PersianWebDocumentRetrieval\citep{10553090}}
This dataset consists of queries collected from the Zarebin search engine and human-labeled documents, specifically curated for the task of information retrieval from the web.

\textbf{NeuCLIR2023Retrieval~\citep{Lawrie2023OverviewOT} and NeuCLIR2022Retrieval~\citep{lawrie2024overviewtrec2023neuclir}} 
The NUECLIR dataset is designed for the Cross-Language Information Retrieval (CLIR) task, where systems process queries in one language (English) and retrieve relevant news articles written in a different language, such as Chinese, Persian, or Russian. The Persian collection within this dataset includes Persian documents and corresponding English queries, which are scheduled for release. The query object comprises both human-translated and machine-translated queries, facilitating monolingual retrieval and cross-language retrieval scenarios.

\textbf{WikipediaRetrievalgMultilingual~\citep{wikidump}} 
The dataset utilized in this study is derived from Cohere's Wikipedia-2023-11 dataset, which comprises a comprehensive collection of Wikipedia articles. Additionally, the dataset includes synthetically generated queries designed to facilitate information retrieval tasks.

\textbf{MIRACL~\citep{zhang-etal-2023-miracl}} The MIRACL dataset is a multilingual dataset for information retrieval. It comprises documents from Wikipedia and human-generated queries corresponding to these documents. One of the languages supported by this dataset is Persian, which we utilize for the information retrieval task.

\subsection{Reranking}
\textbf{MIRACLReranking} MIRACL-Reranking is a subset of the MIRACL dataset, where, for each query, a collection of documents is retrieved and annotated as relevant or irrelevant.

\textbf{WikipediaRerankingMultilingual~\citep{wikidump}} 
The dataset is sourced from Cohere's Wikipedia-2023-11 dataset and includes synthetically generated queries.

\subsection{Pair Classification}
\textbf{CExaPPC~\citep{9786243}}
ExaPPC is an extensive paraphrase corpus consisting of monolingual Persian sentence-level paraphrases, derived from a variety of sources.

\textbf{SynPerChatbotRAGFAQPC}
This dataset is derived from the \textit{Synthetic Persian Chatbot RAG Dataset}.  
It evaluates the model's ability to identify relevant FAQ question-answer pairs corresponding to the user's new message in a chatbot conversation, considering the message history within a RAG-based system.

\textbf{FarsiParaphraseDetection~\citep{farsi_paraphrase_detection}} This dataset consists of 7,826 pairs of sentences in Persian, manually annotated for paraphrase detection.

\textbf{FarsTail~\citep{DBLP:journals/corr/abs-2009-08820}} FarsTail dataset is a Persian textual entailment dataset designed to facilitate natural language understanding tasks.

\textbf{SynPerTextKeywordsPC}
This dataset is derived from the \textit{Synthetic Persian Keywords Dataset}.  
It is designed to evaluate the model's ability to identify the keywords of a given text.

\textbf{ParsinluEntail~\citep{khashabi-etal-2021-parsinlu}} 
The dataset focuses on a Persian textual entailment task, which involves determining whether one sentence entails another sentence. The questions are partially derived from translations of the SNLI dataset and partially generated by expert annotators.

\textbf{ParsinluQueryParaphPC~\citep{khashabi-etal-2021-parsinlu}} This study addresses a Persian query paraphrasing task, which involves determining whether two questions are paraphrases of each other. The dataset comprises questions that are partially generated using Google’s auto-complete feature and partially translated from the Quora Paraphrasing dataset.

\textbf{SynPerQAPC}
This dataset is derived from the \textit{Synthetic Persian QA Dataset}.  
It is designed to evaluate the model's ability to identify the correct answers for given questions.

\begin{table*}[h!]
\small
\centering
\begin{tabular}{l|ccccc}

\textbf{Datasets} & Type & Language & \#Train & \#Dev. & \#Test \\
\hline
        DigikalamagClassification & classification & Persian & 6896 & 767 & 852 \\
        NLPTwitterAnalysisClassification & classification & Persian & 2715 & 1357 & 1360 \\
        SentimentDKSF & classification & Persian & 28602 & 0 & 2315 \\
        PersianTextEmotion & classification & Persian & 5558 & 0 & 1390 \\
        PersianFoodSentimentClassification & classification & Persian & 56700 & 6300 & 7000 \\
        DeepSentiPers & classification & Persian & 6320 & 703 & 1854 \\
        MassiveIntentClassification & classification & Multilingual & 11514 & 2033 & 2974 \\
        MassiveScenarioClassification & classification & Multilingual & 11514 & 2033 & 2974 \\
        SynPerChatbotConvSAClassification & classification & Persian & 4496 & 0 & 1499 \\
        SynPerChatbotToneUserClassification & classification & Persian & 8709 & 0 & 1537 \\
        SynPerChatbotToneChatbotClassification & classification & Persian & 8709 & 0 & 1537 \\
        SynPerChatbotRAGToneUserClassification & classification & Persian & 3261 & 0 & 1087 \\
        SynPerChatbotRAGToneChatbotClassification & classification & Persian & 3261 & 0 & 1087 \\
        SynPerChatbotConvSAToneUserClassification & classification & Persian & 4496 & 0 & 1499 \\
        SynPerChatbotConvSAToneChatbotClassification & classification & Persian & 4496 & 0 & 1499 \\
        SynPerChatbotSatisfactionLevelClassification & classification & Persian & 8709 & 0 & 1537 \\
        SynPerTextToneClassification & classification & Persian & 16587 & 0 & 2928 \\
        SIDClassification & classification & Persian & 8712 & 0 & 3735 \\
        \hline
        HamshahriClustring & clustering & Persian & 0 & 0 & 2203 \\
        DigikalamagClustering & clustering & Persian & 6896 & 767 & 852 \\
        NLPTwitterAnalysisClustering & clustering & Persian & 2715 & 1357 & 1360 \\
        BeytooteClustering & clustering & Persian & 0 & 0 & 95851 \\
        SIDClustring & clustering & Persian & 8712 & 0 & 3735 \\
        \hline
        Farsick & STS & Persian & 0 & 0 & 8566 \\
        SynPerSTS & STS & Persian & 70155 & 0 & 12385 \\
        Query2Query & STS & Persian & 4228933 & 0 & 42717 \\
        \hline
        CExaPPC & pair classification & Persian & 63021 & 13505 & 13504 \\
        FarsiParaphraseDetection & pair classification & Persian & 6260 & 783 & 783 \\
        ParsinluEntail & pair classification & Persian & 755 & 270 & 1675 \\
        ParsinluQueryParaphPC & pair classification & Persian & 1830 & 898 & 1916 \\
        FarsTail & pair classification & Persian & 0 & 0 & 1564 \\
        SynPerChatbotRAGFAQPC & pair classification & Persian & 6522 & 0 & 2174 \\
        SynPerTextKeywordsPC & pair classification & Persian & 33174 & 0 & 5856 \\
        SynPerQAPC & pair classification & Persian & 500106 & 0 & 55568 \\
        \hline
        PersianWebDocumentRetrieval & retrieval & Persian & 245692 & 0 & 175472 \\
        NeuCLIR2023Retrieval & retrieval & Multilingual & 0 & 0 & 2258678 \\
        NeuCLIR2022Retrieval & retrieval & Multilingual & 0 & 0 & 2266190 \\
        WikipediaRetrievalMultilingual & retrieval & Multilingual & 0 & 0 & 15000 \\
        MIRACL & retrieval & Multilingual & 0 & 0 & 2213743 \\
        SynPerChatbotRAGFAQRetrieval & retrieval & Persian & 11957 & 0 & 9783 \\
        SynPerChatbotTopicsRetrieval & retrieval & Persian & 34084 & 0 & 11128 \\
        SynPerChatbotRAGTopicsRetrieval & retrieval & Persian & 14590 & 0 & 7648 \\
        SynPerQARetrieval & retrieval & Persian & 520695 & 0 & 298426 \\
        ArguAna-Fa & retrieval & Persian & 0 & 0 & 10080 \\
        ClimateFEVER-Fa & retrieval & Persian & 0 & 0 & 5421274 \\
        CQADupstack-Fa & retrieval & Persian & 0 & 0 & 480902 \\
        DBPedia-Fa & retrieval & Persian & 0 & 0 & 4651208 \\
        FEVER-Fa & retrieval & Persian & 5556643 & 0 & 5424495 \\
        FIQA2018-Fa & retrieval & Persian & 71804 & 0 & 59344 \\
        HotpotQA-Fa & retrieval & Persian & 5403329 & 0 & 5248139 \\
        MSMARCO-Fa & retrieval & Persian & 9374574 & 0 & 8845925 \\
        NFCorpus-Fa & retrieval & Persian & 114208 & 0 & 15967 \\
        NQ-Fa & retrieval & Persian & 0 & 0 & 2685669 \\
        Quora-Fa & retrieval & Persian & 0 & 0 & 538606 \\
        SCIDOCS-Fa & retrieval & Persian & 0 & 0 & 30585 \\
        SciFact-Fa & retrieval & Persian & 6102 & 0 & 5522 \\
        Touche2020-Fa & retrieval & Persian & 0 & 0 & 383477 \\
        TRECCOVID-Fa & retrieval & Persian & 0 & 0 & 196005 \\
        \hline
        MIRACLReranking & reranking & Multilingual & 0 & 0 & 1314 \\
        WikipediaRerankingMultilingual & reranking & Multilingual & 0 & 0 & 1500 \\
        \hline
        SynPerChatbotSumSRetrieval & summary retrieval & Persian & 8709 & 0 & 1537 \\
        SynPerChatbotRAGSumSRetrieval & summary retrieval & Persian & 3261 & 0 & 1087 \\
        SAMSumFa & summary retrieval & Persian & 14045 & 0 & 1561 \\

\end{tabular}
\caption{The number of documents in each of the FaMTEB datasets.}
\label{tab:faMTEB_count}
\end{table*}

\begin{table*}[h!]
\small
\centering
\begin{tabular}{c|ccc}

\textbf{} & \textbf{Synthetic Persian STS} & \textbf{Synthetic Persian Keywords / Tone} & \textbf{type} \\
\hline
\href{https://www.wikipedia.org/}{wikipedia} & 5000 & 1000 & wiki \\
\href{https://wikishia.net/}{wikishia} & 500 & 500 & wiki \\
\href{https://wiki.ahlolbait.com/}{wiki.ahlolbait} & 500 & 500 & wiki \\
\href{https://hawzah.net/}{hawzah} & 500 & 500 & wiki \\
\hline
\href{https://www.yjc.ir/}{yjc} & 1000 & 1000 & news \\
\href{https://www.tasnimnews.com/}{tasnim} & 500 & 1000 & news \\
\href{https://www.tabnak.ir/}{tabnak} & 500 & 1000 & news \\
\href{https://www.beytoote.com/}{beytoote} & 500 & 1000 & news \\
\href{https://www.varzesh3.com/}{varzesh3} & 500 & 1000 & news \\
\href{https://www.isna.ir/}{isna} & 1000 & 1000 & news \\
\href{https://www.mehrnews.com/}{mehrnews} & 1000 & 1000 & news \\
\href{https://www.asriran.com/}{asriran} & 1000 & 1000 & news \\
\href{https://www.khabaronline.ir/}{khabaronline} & 500 & 500 & news \\
\href{https://ecoiran.com/}{ecoiran} & 500 & 500 & news \\
\href{https://www.hamshahrionline.ir/}{hamshahrionline} & 0 & 500 & news \\
\href{https://donya-e-eqtesad.com/}{donyaeeqtesad} & 0 & 500 & news \\
\hline
\href{https://www.alibaba.ir/}{alibaba} & 500 & 500 & article \\
\href{https://khanesarmaye.com/}{khanesarmaye} & 500 & 0 & article \\
\href{https://digiato.com/}{digiato} & 500 & 500 & article \\
\href{https://www.ninisite.com/article}{ninisite / article} & 500 & 500 & article \\
\href{https://www.zoomit.ir/}{zoomit} & 500 & 500 & article \\
\href{https://bigbangpage.com/}{bigbangpage} & 500 & 500 & article \\
\href{https://namnak.com/}{namnak} & 500 & 500 & article \\
\href{https://hamgardi.com/}{hamgardi} & 0 & 500 & article \\
\hline
\href{https://www.ninisite.com/discussion}{ninisite / discussion} & 500 & 2000 & informal \\
\href{https://www.voolak.com/}{voolak} & 500 & 1500 & informal \\
\hline
\href{https://doctoreto.com/}{doctoreto} & 500 & 500 & others \\
\href{https://taaghche.com/}{taaghche} & 500 & 500 & others \\
\href{https://virgool.io/}{virgool} & 500 & 0 & others \\
\href{https://soft98.ir/}{soft98} & 0 & 500 & others \\
\href{https://vipofilm.com/}{vipofilm} & 0 & 500 & others \\

\end{tabular}
\caption{The number of documents selected for each dataset from various websites.}
\label{tab:doc_num}
\end{table*}

\begin{table*}[ht]
    \centering
    \begin{adjustbox}{width=\textwidth}
    \begin{tabular}{ccccccccccccccccccc}
        \hline
         car-news & news & computer & attire & health & fun & cookery & art & iran & psychology & baby & sport & wedlock & religious & scientific & mode & housekeeping & pictures & job \\
         \hline
         12669 & 12000 & 8603 & 6109 & 5333 & 5241 & 4869 & 4741 & 4441 & 4239 & 4107 & 3889 & 3856 & 3700 & 3578 & 3047 & 2951 & 1251 & 1227 \\
         \hline
    \end{tabular}
    \end{adjustbox}
    \caption{The number of samples in each category of dataset BeytooteClustering.}
    \label{tab:beytoote}
\end{table*}

\section{Data Evaluation}
\label{sec:data_evaluation}
\subsection{Sythetic Data}
In this section, we evaluate the synthetic datasets that were generated. To do so, we randomly selected a number of samples from each synthetic dataset for manual tagging by human annotators. Based on these tags, we calculated a metric for each dataset that reflects the accuracy of the labels generated by the LLM model. The results are presented in Table \ref{tab:syn_data_eval}.

It is important to note that for the STS datasets and the SynPerChatbotSatisfactionLevelClassification dataset, where the labels are ordinal, we used the correlation coefficient (measuring the correlation between the human annotator labels and the labels generated by the LLM) as the evaluation metric. For the other datasets, we used accuracy as the evaluation metric.

Additionally, for retrieval datasets, each sample for tagging consisted of a pair: a query and a related document. The calculated accuracy indicates the percentage of samples where the query and document are indeed related.

For classification, pair classification, and STS datasets, a number of samples were considered for each class. The annotators determined which class each sample truly belongs to, and the metric was calculated based on these annotations.

Another noteworthy point is that each tagging sample was labeled by two annotators. For STS datasets and SynPerChatbotSatisfactionLevelClassification dataset (where the evaluation metric is the correlation coefficient), the final label is the average of the two annotators' labels. For other datasets (where accuracy is the metric), if the two annotators' labels for a sample were identical, the final label for that sample was the same. If the annotators' labels differed, a third annotator determined the final label.

Furthermore, since we evaluated the Query2Query dataset similarly to synthetic STS datasets, its evaluation results are also included in this section and Table \ref{tab:syn_data_eval}, even though this dataset was not generated by the LLM.

As shown in Table \ref{tab:syn_data_eval}, all datasets exhibit relatively acceptable accuracy, with most datasets achieving an accuracy of over 90

\begin{table*}[h!]
\small
\centering
\begin{tabular}{l|ccc}

\textbf{Dataset} & \textbf{Evaluation value} & \textbf{Evaluation metric} & \textbf{Dataset type} \\
\hline
SynPerChatbotConvSAClassification & 93\% & Accuracy & Classification \\
SynPerChatbotToneUserClassification & 79\% & Accuracy & Classification \\
SynPerChatbotToneChatbotClassification & 89\% & Accuracy & Classification \\
SynPerChatbotRAGToneUserClassification & 85\% & Accuracy & Classification \\
SynPerChatbotRAGToneChatbotClassification & 86\% & Accuracy & Classification \\
SynPerChatbotConvSAToneUserClassification & 94\% & Accuracy & Classification \\
SynPerChatbotConvSAToneChatbotClassification & 94\% & Accuracy & Classification \\
SynPerChatbotSatisfactionLevelClassification & 0.90 & Corr. coef. & Classification \\
SynPerTextToneClassification & 76.0\% & Accuracy & Classification \\
\hline
SynPerChatbotSumSRetrieval & 100\% & Accuracy & Summary Retrieval \\
SynPerChatbotRAGSumSRetrieval & 99\% & Accuracy & Summary Retrieval \\
\hline
SynPerChatbotRAGFAQRetrieval & 77\% & Accuracy & Retrieval \\
SynPerChatbotTopicsRetrieval & 88.5\% & Accuracy & Retrieval \\
SynPerChatbotRAGTopicsRetrieval & 93.0\% & Accuracy & Retrieval \\
SynPerQARetrieval & 98\% & Accuracy & Retrieval \\
\hline
SynPerChatbotRAGFAQPC & 85.5\% & Accuracy & Pair Classification \\
SynPerTextKeywordsPC & 94.5\% & Accuracy & Pair Classification \\
SynPerQAPC & 95.0\% & Accuracy & Pair Classification \\
\hline
SynPerSTS & 0.911 & Corr. coef. & STS \\
Query2Query & 0.695 & Corr. coef. & STS \\

\end{tabular}
\caption{Evaluation of synthetic datasets.}
\label{tab:syn_data_eval}
\end{table*}

\subsection{Translate Data}
Inspired by the MMARCO paper~\citep{DBLP:journals/corr/abs-2108-13897} to evaluate translated data, we use the comparison of accuracy measurements in English and Persian using the BM25 metric. In this approach, we index the entire corpus of each dataset using ElasticSearch and retrieve test queries using the BM25 metric. Finally, we report the recall of each dataset for both languages. In Table \ref{tab:translate_data_eval}, we observe that the information retrieval accuracy in the translated data does not differ significantly from the accuracy in English, and it can be concluded that the BEIR translated data meets the necessary quality standards.

\newcolumntype{R}[2]{%
    >{\adjustbox{angle=#1,lap=\width-(#2)}\bgroup}%
    c%
    <{\egroup}%
}
\newcommand*\rot{\multicolumn{1}{R{90}{1em}}}
\begin{table*}[h!]

\centering
\begin{adjustbox}{width=\textwidth}
\begin{tabular}{c|c|cccccccccccccccc}
\textbf{Language} & Evaluation metric & \rot{ArguAna} & \rot{ClimateFEVER} & \rot{CQADupstack} & \rot{DBPedia} & \rot{FEVER} & \rot{FIQA} & \rot{HotpotQA} & \rot{MSMARCO} & \rot{NFCorpus} & \rot{NQ} & \rot{Quora} & \rot{SCIDOCS} & \rot{SciFact} & \rot{Touche2020} & \rot{TRECCOVID} \\
\hline
Persian & ndcg@10 & 42.93 & 15.32 & 26.58 & 25.45 & 50.84 & 17.44 & 46.53 & 38.27 & 33.25 & 22.02 & 61.71 & 14.06 & 65.31 & 26.49 & 56.9 \\
English & ndcg@10 & 47.16 & 18.61 & 32.52 & 32.01 & 64.93 & 25.36 & 60.22 & 47.68 & 34.28 & 32.60 & 88.77 & 16.46 & 69.06 & 34.70 & 68.80 \\
\hline
Persian & recall@100 & 93.67 & 34.29 & 54.63 & 35.83 & 82.98 & 46.94 & 65.11 & 33.02 & 26.50 & 60.78 & 86.03 & 32.31 & 89.7 & 47.18 & 10.50 \\
English & recall@100 & 95.16 & 40.93 & 62.13 & 43.47 & 92.15 & 54.88 & 76.30 & 45.02 & 26.02 & 78.28 & 97.69 & 36.75 & 91.92 & 56.09 & 11.73 \\

\end{tabular}
\end{adjustbox}
\caption{Accuracy evaluations conducted on English and Persian languages using the BM25 metric.}
\label{tab:translate_data_eval}
\end{table*}

\subsection{Model Accuracy}
The accuracy of each model, presented separately, is provided in Table \ref{tab:full_accuracy_table}.

\begin{table*}[h!]

\centering
\begin{adjustbox}{width=\textwidth}
\begin{tabular}{c ccccccccccccccc}
 & \rot{sentence-transformer-parsbert-fa} & \rot{RoBERTa-WLNI} & \rot{BERT-WLNI} & \rot{faBert} & \rot{ParsBERT} & \rot{paraphrase-multilingual-MiniLM-L12-v2} & \rot{LaBSE} & \rot{TookaBERT-Base} & \rot{Tooka-SBERT} & \rot{GTE-multilingual-base} & \rot{multilingual-e5-base} & \rot{multilingual-e5-large} & \rot{BGE-m3-unsupervised} & \rot{BGE-m3} & \rot{Jina-embeddings-v3} \\
\hline
    \textit{Classification} \\
\hline
    PersianFoodSentimentClassification & 64.25 & 64.52 & 64.17 & 70.3 & 73.75 & 73.46 & 72.09 & 72.01 & 80.05 & 77.49 & 75.05 & 76.31 & 77.34 & 83.4 & 83.57  \\ 
    SynPerChatbotConvSAClassification & 62.17 & 59.44 & 58.23 & 70.51 & 77.1 & 57.51 & 75.41 & 78.07 & 76.38 & 63.29 & 64.61 & 60.77 & 63.15 & 61.03 & 71.57  \\ 
    SynPerChatbotConvSAToneChatbotClassification & 70.01 & 68 & 65.39 & 88.43 & 91.21 & 56.6 & 66.77 & 88.97 & 60.75 & 49.09 & 63.18 & 58.07 & 54.42 & 50.55 & 51.88  \\ 
    SynPerChatbotConvSAToneUserClassification & 55.72 & 62.01 & 61.77 & 48.36 & 72.68 & 50.35 & 53.63 & 69.6 & 56.46 & 51.86 & 48.85 & 52.6 & 51.3 & 48.85 & 52.86  \\ 
    SynPerChatbotSatisfactionLevelClassification & 26.14 & 21.98 & 23.05 & 31.87 & 35.6 & 22.04 & 35.02 & 37.22 & 37.18 & 30.82 & 25.32 & 25.23 & 26.04 & 24.72 & 35.43  \\ 
    SynPerChatbotRAGToneChatbotClassification & 38.65 & 43.02 & 42.41 & 58.41 & 61.39 & 38.41 & 42.97 & 58.31 & 38.69 & 32.41 & 35.15 & 37.16 & 35.16 & 35.45 & 33.16  \\ 
    SynPerChatbotRAGToneUserClassification & 47.07 & 58.03 & 53.84 & 50.9 & 62.37 & 44.68 & 54 & 62.59 & 51.32 & 48.45 & 44.9 & 50.73 & 50.7 & 48.47 & 49  \\ 
    SynPerChatbotToneChatbotClassification & 48.73 & 48.99 & 48.26 & 67.53 & 74.99 & 41.14 & 52.42 & 72.46 & 47.68 & 33.64 & 42.36 & 41.5 & 42.38 & 37.92 & 36.23  \\ 
    SynPerChatbotToneUserClassification & 44.93 & 56.96 & 52.97 & 41.76 & 60.17 & 41.67 & 51.15 & 57.25 & 46.91 & 43.2 & 39.98 & 46.73 & 45.75 & 42.71 & 45.18  \\ 
    PersianTextTone & 55.86 & 71.12 & 73.12 & 91.3 & 89.72 & 46.39 & 58.71 & 89.78 & 51.8 & 51.7 & 63.69 & 70.19 & 61.54 & 55.67 & 50.69  \\ 
    SIDClassification & 55 & 50.41 & 48.26 & 54 & 55.7 & 54.46 & 56.71 & 58.75 & 53.24 & 60.62 & 60.73 & 61.37 & 60.32 & 59.62 & 61.68  \\ 
    DeepSentiPers & 42.51 & 43.38 & 44.28 & 42.41 & 50.63 & 55.91 & 60.91 & 54.53 & 64.43 & 57.95 & 61.9 & 60.95 & 58.89 & 67.51 & 65.26  \\ 
    PersianTextEmotion & 38.81 & 37.59 & 37.5 & 48.8 & 48.18 & 45.37 & 53.45 & 53.26 & 57.01 & 51.5 & 54.85 & 61.88 & 58.73 & 61.13 & 51.88  \\ 
    SentimentDKSF & 49.17 & 49.87 & 51.52 & 59.04 & 59.69 & 65.62 & 67.67 & 58.16 & 69.57 & 67.52 & 71.26 & 71.07 & 66.34 & 75.35 & 75.15  \\ 
    NLPTwitterAnalysisClassification & 73.68 & 70.54 & 70.74 & 70.29 & 70.71 & 74.98 & 74.93 & 70.28 & 74.72 & 75.62 & 74.67 & 75.99 & 77.75 & 76.93 & 75.97  \\ 
    DigikalamagClassification & 79.74 & 74.79 & 74.58 & 82.59 & 81.8 & 74.66 & 85.12 & 82.37 & 72.95 & 82.93 & 86.78 & 87.05 & 86.31 & 86.03 & 84.88  \\ 
    MassiveIntentClassification (fa) & 44.19 & 51.43 & 52.78 & 55.09 & 60.07 & 61.03 & 62.33 & 59.98 & 63.73 & 62.29 & 59.51 & 63.74 & 66.17 & 69.44 & 72.6  \\ 
    MassiveScenarioClassification (fa) & 51.78 & 59.53 & 58.24 & 58.46 & 62.6 & 65.89 & 67.43 & 64.11 & 67.45 & 67.88 & 63.92 & 67.55 & 72.37 & 73.29 & 81.78  \\
\hline
\textit{Clustering} \\
\hline
    BeytooteClustering & 61.95 & 61.92 & 60.75 & 53.81 & 51.6 & 63 & 56.12 & 55.27 & 55.62 & 62.52 & 59.16 & 61.5 & 61.44 & 60.71 & 63.4  \\
    DigikalamagClustering & 60.53 & 46.24 & 55.82 & 38.71 & 50.73 & 48.69 & 44.07 & 45.55 & 41.03 & 34.39 & 38.63 & 39.89 & 47.48 & 39.56 & 43.3  \\
    HamshahriClustring & 75.57 & 68.85 & 67.37 & 65.08 & 64.84 & 63.84 & 67.09 & 66.43 & 63.28 & 69.83 & 67.83 & 67.42 & 67.55 & 69.48 & 66.88  \\
    NLPTwitterAnalysisClustering & 79.18 & 76.24 & 77.85 & 74.55 & 74.46 & 78.97 & 76.12 & 70.06 & 82.24 & 80.82 & 78.18 & 78.48 & 80.63 & 80.9 & 80.69  \\
    SIDClustring & 46.91 & 39.8 & 39.79 & 43.77 & 39.12 & 36.11 & 39.4 & 41.3 & 40.08 & 38.86 & 38.79 & 38.65 & 41.02 & 38 & 41.5  \\
\hline
\textit{Pair Classification} \\
\hline
FarsTail & 58.92 & 54.98 & 56.09 & 57.15 & 57.79 & 64.84 & 62.93 & 55.6 & 81.52 & 72.65 & 70.76 & 69.74 & 69.77 & 73.14 & 71.85  \\
        CExaPPC & 82.72 & 93.7 & 94.29 & 90.88 & 91.13 & 95.95 & 98.97 & 94.55 & 98.8 & 98.42 & 98.7 & 98.97 & 97.26 & 99.09 & 97.41  \\
        SynPerChatbotRAGFAQPC & 54.02 & 55.81 & 52.54 & 58.26 & 50.84 & 60.75 & 62.68 & 60.32 & 71.32 & 66.77 & 65.42 & 62.9 & 65.03 & 64.43 & 62.06  \\
        FarsiParaphraseDetection & 90.38 & 92.02 & 93.01 & 77.99 & 93.69 & 96.82 & 94.34 & 86.98 & 95.99 & 97.06 & 96.39 & 97.57 & 96.86 & 95.57 & 96.6  \\
        SynPerTextKeywordsPC & 86.94 & 82.59 & 81.97 & 76.02 & 71.25 & 89.93 & 87.88 & 77.68 & 94.93 & 96.4 & 95.73 & 94.79 & 96.21 & 97.04 & 94.96  \\
        SynPerQAFaPC & 65.13 & 73.11 & 73.52 & 66.6 & 67.84 & 80.6 & 83.07 & 66.2 & 86.34 & 91.66 & 94.24 & 95.16 & 93.2 & 93.76 & 93.38  \\
        ParsinluEntail & 55.14 & 55.9 & 55.58 & 57.28 & 57.79 & 69.19 & 60.76 & 56.39 & 77.7 & 66.55 & 64.81 & 65.43 & 59.81 & 68.65 & 65.82  \\
        ParsinluQueryParaphPC & 71.18 & 60.72 & 61.57 & 65.67 & 62.8 & 81.15 & 80.4 & 67.83 & 89.7 & 87.09 & 86.3 & 87.75 & 86.41 & 89.98 & 87.63  \\
\hline
\textit{Reranking} \\
\hline
    MIRACLReranking (fa) & 18.34 & 17.38 & 18.17 & 23.72 & 18.9 & 30.83 & 29.05 & 14.4 & 35.87 & 55.05 & 57.36 & 59.36 & 48.78 & 60.92 & 42.91  \\
    WikipediaRerankingMultilingual (fa) & 61.47 & 72.11 & 73.28 & 77.43 & 73.86 & 80.8 & 82.42 & 73.95 & 80.71 & 84.38 & 86.78 & 89.32 & 90.71 & 88.21 & 79.61 \\
\hline
\textit{Retrieval} \\
\hline
    SynPerQARetrieval & 20.36 & 26.59 & 25.02 & 42.94 & 24.95 & 52.45 & 53.99 & 26.88 & 65.02 & 77.44 & 85.59 & 87.35 & 85.14 & 86.27 & 85.4  \\
    SynPerChatbotTopicsRetrieval & 2.38 & 3.52 & 4.23 & 0.05 & 0.15 & 12.28 & 6.2 & 0.16 & 10.76 & 28.07 & 15.37 & 11.82 & 10.59 & 19.18 & 18.75  \\
    SynPerChatbotRAGTopicsRetrieval & 4.33 & 4.59 & 4.72 & 0.09 & 1.19 & 16.39 & 12.1 & 0.45 & 18.93 & 30.97 & 20.11 & 19.24 & 13.22 & 19.91 & 24.26  \\
    SynPerChatbotRAGFAQRetrieval & 6.76 & 7.46 & 6.37 & 10.02 & 5.09 & 19.22 & 18.82 & 12.24 & 24.3 & 31.47 & 28.49 & 23.48 & 30.84 & 32.04 & 47.46  \\
    PersianWebDocumentRetrieval & 12.61 & 8.14 & 12.55 & 7.95 & 10.04 & 14.31 & 28.21 & 10.85 & 43.9 & 44.15 & 46.72 & 46.76 & 38.18 & 44.09 & 40.32  \\
    NeuCLIR2022Retrieval & 1.33 & 3.9 & 3.23 & 2.92 & 0.38 & 19.78 & 2.56 & 0.02 & 26.96 & 36.67 & 9.75 & 5.3 & 12.12 & 15.48 & 18.25 \\
    NeuCLIR2023Retrieval & 6.6 & 5.27 & 5.02 & 12.1 & 1.86 & 26.34 & 21.52 & 4.63 & 36.47 & 50.93 & 46.1 & 46.67 & 46.53 & 52.2 & 51.45 \\
    WikipediaRetrievalMultilingual (fa) & 35.63 & 37.34 & 41.29 & 63.67 & 48.41 & 62.15 & 67.06 & 46.14 & 79.02 & 84.94 & 88.11 & 90.4 & 91.19 & 89.32 & 89.02  \\
    MIRACLRetrieval (fa) & 1.95 & 4.34 & 4.35 & 8.24 & 4.52 & 13.33 & 10.53 & 2.21 & 21.32 & 53.89 & 57.48 & 59.01 & 39.93 & 60.9 & 55.21  \\
    ClimateFEVER-Fa & 1.13 & 2.68 & 2.15 & 5.06 & 2.05 & 12.23 & 3.73 & 0.39 & 9.47 & 18.83 & 12.6 & 12.75 & 16.41 & 24.31 & 29.87  \\
    FEVER-Fa & 0.7 & 1.11 & 1.42 & 1.7 & 0.59 & 18 & 7 & 0.41 & 8.44 & 61.33 & 48.05 & 41.56 & 44.74 & 55.99 & 63.75 \\
    DBPedia-Fa & 1.92 & 1.13 & 2.19 & 2.87 & 1.92 & 11.53 & 10.78 & 1.2 & 13.77 & 29.2 & 28.74 & 30.36 & 22.47 & 29.85 & 31.84  \\
    HotpotQA-Fa & 0.22 & 0.85 & 0.85 & 6.52 & 3.37 & 12.39 & 11.94 & 2.33 & 16.44 & 49.04 & 55.33 & 60.15 & 39.24 & 56.54 & 51.43  \\
    MSMARCO-Fa & 1.04 & 1.5 & 1.34 & 2.02 & 1.14 & 7.89 & 6.43 & 1.23 & 9.33 & 23.33 & 26.88 & 30.92 & 21.38 & 29.09 & 29.85  \\
    NQ-Fa & 0.62 & 1.88 & 1.75 & 3.49 & 1.27 & 11.49 & 7.88 & 1.08 & 11.97 & 38.8 & 39.84 & 44.82 & 26.69 & 46.62 & 50.33  \\
    ArguAna-Fa & 20.59 & 18.77 & 14.94 & 22.24 & 21.8 & 36.45 & 36.13 & 27.51 & 31.88 & 50.4 & 43.19 & 45.5 & 55.28 & 50.4 & 34.88  \\
    CQADupstackRetrieval-Fa & 2.46 & 4.89 & 4.45 & 9.46 & 4.49 & 18.35 & 16.65 & 3.09 & 18.09 & 26.03 & 29.87 & 31.59 & 32.45 & 31.72 & 27.91  \\
    FiQA2018-Fa & 0.87 & 2.61 & 2.22 & 3.1 & 1.58 & 10.31 & 6.35 & 1.82 & 11.22 & 26.47 & 23.17 & 30.15 & 27.39 & 30.38 & 34.43  \\
    NFCorpus-Fa & 5.36 & 3.45 & 4 & 7.14 & 3.36 & 14.83 & 15.52 & 4.86 & 19.7 & 25.61 & 25.47 & 28.59 & 28.28 & 29.47 & 28.21  \\
    QuoraRetrieval-Fa & 47.21 & 46.38 & 47.38 & 52.45 & 49.29 & 73.61 & 72.52 & 51.72 & 76.67 & 77.99 & 77.26 & 79.96 & 80.14 & 82.18 & 59.44  \\
    SCIDOCS-Fa & 1.85 & 2.94 & 1.66 & 3.74 & 1.56 & 9.28 & 5.8 & 1.97 & 9.07 & 12.72 & 11.76 & 11.58 & 13.55 & 14.56 & 14.52  \\
    SciFact-Fa & 5.95 & 6.48 & 7.75 & 18.86 & 10.33 & 31.54 & 34.57 & 13.54 & 37.19 & 56.15 & 57.79 & 59.69 & 57.76 & 60.52 & 61.51  \\
    Touche2020-Fa & 1.44 & 7.04 & 5.46 & 2.14 & 1.14 & 16.47 & 4.59 & 1.42 & 13.22 & 24.69 & 22.48 & 26.19 & 16.15 & 22.87 & 26  \\
\hline
\textit{STS} \\
\hline
    Farsick & 50.16 & 48.45 & 50.62 & 49.27 & 53.57 & 66.72 & 64.42 & 52.04 & 69.64 & 70.95 & 69.93 & 70.67 & 67.29 & 71.75 & 76.96  \\
    SynPerSTS & 67.67 & 72.66 & 71.99 & 77.16 & 73.98 & 83.31 & 88.47 & 74.55 & 89.08 & 86.89 & 86.7 & 87.98 & 86.91 & 87.59 & 88.73  \\
    Query2Query & 47.38 & 43.66 & 46.21 & 29.61 & 55.37 & 51.69 & 66.28 & 59.08 & 70.52 & 69.41 & 66.71 & 67.49 & 69.22 & 69.71 & 70.27  \\
\hline
\textit{Summary Retrieval} \\
\hline
    SAMSumFa & 26.2 & 5.49 & 8.65 & 14.45 & 11.23 & 45.04 & 83.74 & 9.31 & 88.19 & 85.5 & 92.97 & 92.42 & 92.47 & 97.88 & 96.89  \\
    SynPerChatbotSumSRetrieval & 3.21 & 1.03 & 1.95 & 3.18 & 1.36 & 17.37 & 18.02 & 3.2 & 32.73 & 36.78 & 25.09 & 27.6 & 36.61 & 32.13 & 36.99  \\
    SynPerChatbotRAGSumSRetrieval & 14.71 & 10.62 & 8.45 & 10.34 & 8.68 & 33.32 & 40.76 & 12.35 & 56.27 & 60.37 & 45.68 & 49.81 & 55.93 & 53.21 & 62.63 \\
\hline
\end{tabular}
\end{adjustbox}
\caption{The accuracy of each model on all datasets, presented separately.}
\label{tab:full_accuracy_table}
\end{table*}

\end{document}